\documentclass[sigconf, screen]{acmart}

\usepackage{graphicx}
\usepackage{balance}

\definecolor{FG}{RGB}{34,139,34}
\definecolor{DY}{RGB}{0,139,34}

\usepackage{colortbl}
\usepackage{xcolor}
\usepackage{multirow}
\usepackage[normalem]{ulem}
\useunder{\uline}{\ul}{}

\usepackage[most]{tcolorbox}

\newcommand{\ps}[1]{\textsc{PEST}}

\newcommand{\sysGPT}[1]{\textsc{GPT-4o}}
\newcommand{\sysHateRed}[1]{\textsc{HatReD}}
\newcommand{\sysHateRedAug}[1]{\textsc{HatReDAug}}
\newcommand{\sysMemeSense}[1]{\textsc{MemeSense}}
\newcommand{\sysMemeCap}[1]{\textsc{MemeCap}}

\newcommand{\sysRG}[1]{\textsc{Rouge}}
\newcommand{\sysSBERT}[1]{\textsc{BertScore}}
\newcommand{\sysCSIM}[1]{\textsc{Semantic Similarity}}

\newcommand{\sysP}[1]{\textsc{PaliGemma}}
\newcommand{\sysI}[1]{\textsc{Intern-VL3}}
\newcommand{\sysPx}[1]{\textsc{Pixtral}}
\newcommand{\sysSL}[1]{\textsc{SigLIP}}

\AtBeginDocument{
  }

\copyrightyear{2026}
\acmYear{2026}
\setcopyright{othergov}
\acmConference[MM '26]{Proceedings of the 35th ACM International Conference on Multimedia}{November 10--14, 2026}{Rio de Janeiro, Brazil}
\acmBooktitle{Proceedings of the 35th ACM International Conference on Multimedia (MM '26), November 10--14, 2026, Rio de Janeiro, Brazil}
\acmDOI{10.1145/3767308.3836309}
\acmISBN{979-8-4007-2213-4/2026/11}

\begin{document}

\title{\ps{}: \underline{P}arameter \underline{E}fficient \underline{St}eering of Blackbox VLMs via\\Agentic Few-shot Alignment for Hateful Meme Moderation}

\author{Naquee Rizwan}
\affiliation{%
  \institution{Indian Institute of Technology}
  \city{Kharagpur}
  \state{West Bengal}
  \country{India}}
\email{nrizwan@kgpian.iitkgp.ac.in}

\author{Subhankar Swain}
\affiliation{%
  \institution{Indian Institute of Technology}
  \city{Kharagpur}
  \state{West Bengal}
  \country{India}}
\email{subhankar.swain25@kgpian.iitkgp.ac.in}

\author{Paramananda Bhaskar}
\affiliation{
  \institution{Indian Institute of Technology}
  \city{Kharagpur}
  \state{West Bengal}
  \country{India}}
\email{pbhaskar@kgpian.iitkgp.ac.in}

\author{Shehryaar Shah Khan}
\authornote{Equal contribution}
\affiliation{
  \institution{Indian Institute of Technology}
  \city{Kharagpur}
  \state{West Bengal}
  \country{India}
}
\email{khanshehryaar705@kgpian.iitkgp.ac.in}

\author{Gagan Aryan}
\authornotemark[1]
\affiliation{
 \institution{Indian Institute of Technology}
 \city{Kanpur}
 \state{Uttar Pradesh}
 \country{India}}
\email{gaganaryan19@gmail.com}

\author{Animesh Mukherjee}
\affiliation{
  \institution{Indian Institute of Technology}
  \city{Kharagpur}
  \state{West Bengal}
  \country{India}}
\email{animeshm@cse.iitkgp.ac.in}

\renewcommand{\shortauthors}{Naquee Rizwan et al.}

\begin{abstract}
In this work, we examine hateful memes from three complementary angles -- how to detect them, how to explain their content and how to intervene them before being posted -- by applying a range of strategies built on top of generative AI models. To the best of our knowledge, explanation and intervention have typically been studied separately from detection, which does not reflect real-world conditions. Further, since curating large annotated datasets for meme moderation is prohibitively expensive, we propose a novel framework -- \ps{} -- that leverages task-specific generative VLMs and the few-shot adaptability of large VLMs to cater to different types of memes. We believe this is the first work focused on generalizable hateful meme moderation under limited data conditions, and has strong potential for deployment in real-world production scenarios.
\textit{\textbf{\textcolor{red}{Warning: Contains potentially toxic contents.}}}~\footnote{\textbf{\textcolor{blue}{Accepted to ACM Multimedia 2026 (Main Track)}}}
\end{abstract}

\ccsdesc[500]{Social and professional topics~Hate speech}
\ccsdesc[500]{Information systems~Multimedia and multimodal retrieval}

\keywords{Hateful Memes, Content Moderation, Steering Agents, Few-shot Alignment}

\maketitle

\section{Introduction}
\label{sec:introduction}
Multimodal hate speech research has rapidly grown over the past few years~\cite{hee-etal-2024-recent}, with most works focusing on classification, i.e. assigning memes to labels like hateful, misogynistic, or harmful. This has lead to some famous classification datasets like FHM~\cite{kiela2020hateful} and MAMI~\cite{fersini2022semeval}; there are also some works that have been extended to other languages~\cite{das2023banglaabusememe}. Despite such rapid growth, there are evident research gaps on explaining~\cite{Hee2023DecodingTU,lin2024towards} and intervening~\cite{jha2024memeguard, adak2025memesenseadaptiveincontextframework} these memes, primarily due to the lack of extensive datasets to design, train and evaluate such modular content moderation frameworks. We open-source our dataset and code\footnote{Paper's GitHub repository: \url{https://github.com/hate-alert/PEST}}. 

\begin{figure*}
\centering
  \includegraphics[width=2\columnwidth]{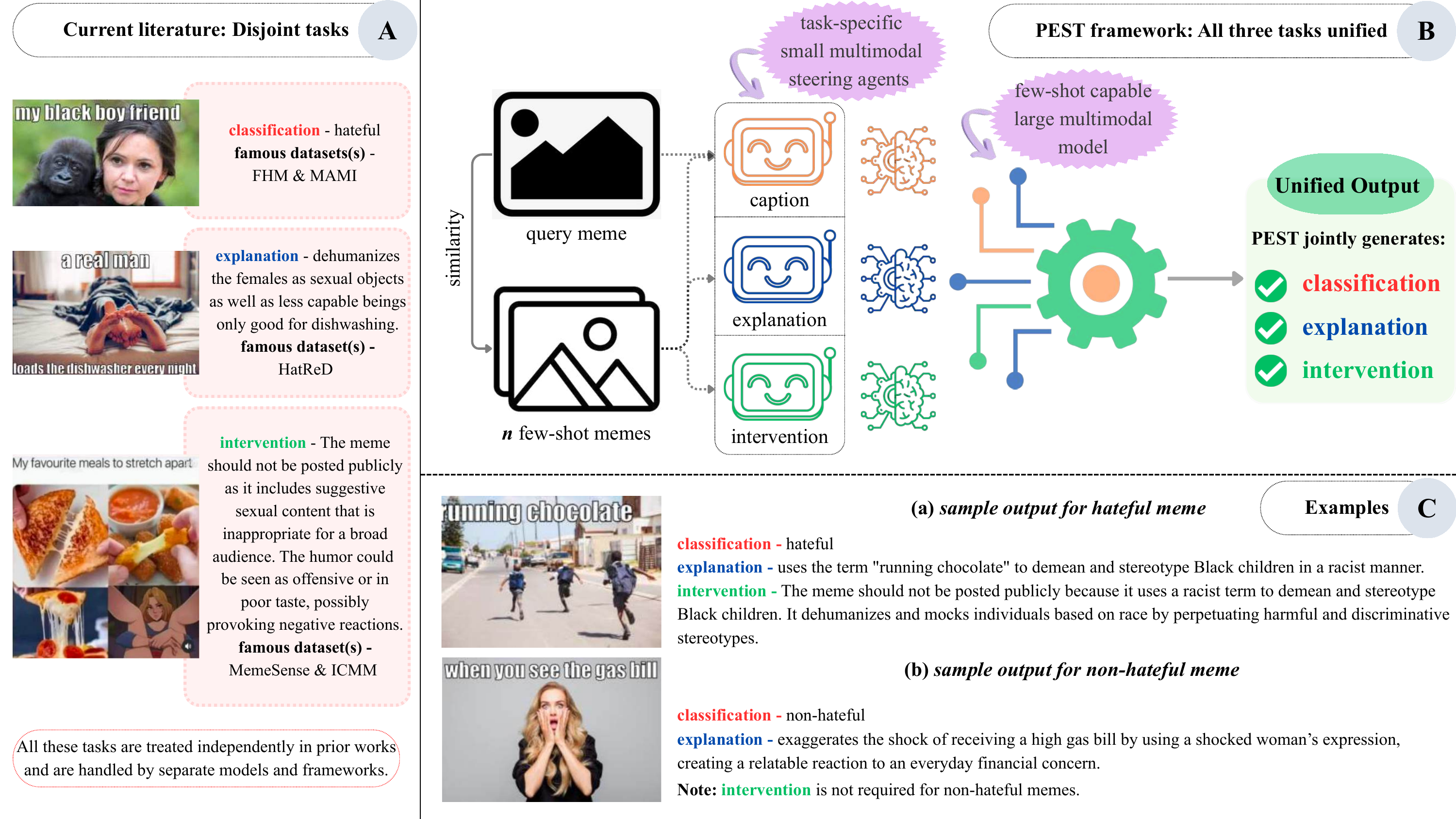}
  \caption{\footnotesize \textbf{A:} Limitations of the current literature,  \textbf{B:} overview of \ps{}-- a \textit{black-box} framework powered by task-specific small agents and few-shot alignment for combined classification, explanation, and intervention of hateful memes, and \textbf{C:} example outputs of unified tasks.}
  \Description{\ps{} framework}
  \label{fig:main_teaser}
\end{figure*}

\noindent\textbf{Difference between explanation and intervention--} As presented in the Figure~\ref{fig:main_teaser}, explanation and intervention are two different tasks and have been introduced separately in the prior works ~\cite{10.1145/3485447.3512260,jha2024memeguard}. Explanation methods characterize \textit{how a model arrives at a prediction by attributing importance to textual, visual, or cross-modal features, thereby improving transparency without altering the input}~\cite{ribeiro2016should,selvaraju2017grad}. In contrast, intervention is \textit{causal and user-facing: it perturbs elements of a meme to assess sensitivity and produces a justification grounded in real-world norms, explaining why the content is inappropriate for sharing}. While prior work in textual hate speech connects intervention to counter speech generation~\cite{Mathew_Saha_Tharad_Rajgaria_Singhania_Maity_Goyal_Mukherjee_2019,Qian_ElSherief_Belding_Wang_2019}, such approaches do not directly extend to hateful memes, where harm arises from multimodal interplay~\cite{Kiela2020HatefulMemes}. Thus, explanation supports interpretability, whereas intervention enables actionable, norm-aware feedback for safer content moderation.

\noindent\textbf{Research gaps--} Research combining the three primary pillars -- classification, explanation, and intervention -- has not been carried out due to the frequent focus of current studies on dealing with each component in isolation~\cite{kmainasi2025memeintel, huang2024towards, agarwal2024mememqa}. Below, we outline the complexity involved in proposing a model capable of performing these three tasks simultaneously. These critical research gaps bolster the difficulty of integrating these pillars.

\noindent \textit{\textbf{(i) Incoherent content moderation of hateful memes --}}  A modular hateful meme moderation framework must perform more than just classifying the meme into specified labels. While there are some works that have explored explanation and intervention, none of them have proposed the simultaneous combination of the three primary complementary pillars. Further in the previous works, explanation and intervention have always been carried when the system is aware of the label (i.e., these systems are evaluated by knowing that the meme is hateful). Logically, this does not resonate with the real-time social media platforms’ content moderation, as these works shadow the importance of classification, simultaneously with explanation and intervention generation. Therefore, to bridge these gaps, we \textbf{formulate a novel task} and a modular framework \ps{} that combines these pillars in a low-resource setup (refer to Figure~\ref{fig:main_teaser} and Section~\ref{sec:experiments}).

\noindent \textit{\textbf{(ii) Lack of coherent datasets --}} Cost and subjectivity associated with the curation of such datasets~\cite{Rizwan_Bhaskar_Das_Majhi_Saha_Mukherjee_2025} has eventually led to no work on these combined pillars to the best of our knowledge. We discuss these dataset limitations in two phases: \textit{(a) training phase}: The dataset proposed in~\cite{Hee2023DecodingTU} (also known as \sysHateRed{} dataset) was the first of its kind to annotate explanations for hateful memes, which is used for training explanation generation frameworks. However, the annotations in this dataset are only for the hateful memes (not for non-hateful memes), thus hindering the development of an end-to-end explainable classification framework. \textit{(b) evaluation phase:} Since we propose this as a new task in the content moderation domain, currently there are no such datasets that can be used to benchmark \ps{} on the three combined pillars. It is very important to assess the proposed modular framework since the data used for training the agents comes from different sources (refer Figures~\ref{fig:main_teaser}, \ref{fig:silver_data} and Sections~\ref{sec:annotation_datasets}, \ref{sec:curated-dataset}).

\noindent Acknowledging these dataset limitations across the two phases, we extend the \textit{test split} of two popular classification datasets (FHM~\cite{kiela2020hateful} and MAMI~\cite{fersini2022semeval}) by annotating them for both explanation and intervention. For training the agents as a part of \ps{}, we also extend \sysHateRed{} to include the explanation of non-hateful memes. We believe that these extensions will be able to new research directions in hateful content moderation.

\noindent \textbf{The low-resource \ps{} framework --} We propose a novel low-resource framework built on top of steering agents to cater for the above-mentioned challenges and limitations. In short, we first fine-tune small VLMs that enact as task-specific agents for silver data generation. Then these silver data aid the larger VLMs with strong few-shot capability to moderate the input through simultaneous classification, explanation, and intervention. For training these agents, we use the prior datasets (\sysMemeCap{}~\cite{hwang2023memecap}, \sysHateRed{}~\cite{Hee2023DecodingTU}, and \sysMemeSense{}~\cite{adak2025memesenseadaptiveincontextframework}) along with the samples we curate in this work (refer Sections~\ref{sec:annotation_datasets} and \ref{sec:curated-dataset}). Finally, we evaluate \ps{} on the extended test split of FHM and MAMI datasets that is also curated in this work. Our approach is grounded in scalable oversight, demonstrating that a small number of lightweight agents ($\sim$3B parameters) can effectively steer much larger vision-language models (VLMs) on complex tasks. The core strength of \ps{} lies in these fully open-source, task-specific agents, which serve as the primary drivers of control. We posit that even models at the scale of \sysGPT{} (reportedly exceeding 200B parameters) can be guided using such compact agents, thereby reinforcing a practical notion of scalability. Importantly, these agents function as parameter-efficient, black-box steering modules that do not require access to or modification of the underlying model weights. In the box below, we provide some of the key results obtained from \ps{}.\\
\noindent
\begin{tcolorbox}[colback=gray!5!white,colframe=gray!75!gray,title=Key results]
$\bullet$~Few-shot classification results using \sysGPT{} attains macro-F1 scores of 80.25\% and 89.07\% on FHM and MAMI datasets respectively, by far outperforming all the existing baselines.\\
$\bullet$~\sysGPT{} outperforms \sysHateRed{} by generating more appropriate explanations for the FHM and MAMI datasets (\sysCSIM{} of $\sim 0.68$ and $\sim 0.65$, respectively).\\
$\bullet$~\sysI{} and \sysPx{} outperform the \sysMemeSense{} framework in terms of intervention generation on FHM and MAMI datasets, respectively (\sysCSIM{} scores of $\sim 0.78$ and $\sim 0.85$, respectively).\\
$\bullet$~\sysGPT{} is the most coherent model when generating both explanation and intervention. Further, across these two tasks, the basic textual properties like \textit{token count}, \textit{ttr}, and \textit{perplexity} are more consistent for \sysGPT{} compared to \sysI{} and \sysPx{}.
\end{tcolorbox}


\section{Related work}
\label{sec:related_works}
\noindent\textbf{Hateful meme moderation --} Most of the prior works on hateful meme moderation have primarily focused on building datasets for hateful meme classification~\cite{kiela2020hateful,fersini2022semeval, lin2024goatbench,swain2025toxictagsdecodingtoxicmemes}. Only a handful of works study explanation~\cite{Hee_Lee_2025,lin2024towards} and intervention~\cite{adak2025memesenseadaptiveincontextframework, jha2024memeguard}. As a result, numerous studies have extensively explored models for classification tasks~\cite{Rizwan_Bhaskar_Das_Majhi_Saha_Mukherjee_2025,cao2023promptingmultimodalhatefulmeme,das2023banglaabusememe}, which has significantly hampered the proposal of frameworks for explanation and interventions~\cite{agarwal2024mememqa,lin2024towards,kmainasi2025memeintel}.\\
\noindent\textbf{Few-shot prompting paradigm --} Training or fine-tuning large generative AI models is costly due to either a huge number of parameters associated with them or due to unavailability of data.
In contrast, few-shot prompting has emerged as a strong paradigm for model alignment without prior fine-tuning and has been often adapted to various different complex tasks where data curation is challenging like in smart agriculture~\cite{PORTO2023100307}, legal systems~\cite{10.1145/3675417.3675513}, robotics~\cite{ayub2022fewshotcontinualactivelearning}, etc. Some works for hateful content detection~\cite{cao2024modularizednetworksfewshothateful,hee-etal-2024-bridging} have also used few-shot prompting. However, to the best of our knowledge, integration of few-shot prompting for simultaneous classification, explanation and intervention generation has not been done so far.

\begin{table}[t]
\centering
\scriptsize
\caption{\footnotesize Statistics of datasets used in our experiments. The upper section details the data used to fine-tune the task-specific small agents (\texttt{paligemma-3b-pt-448}), while the two lower section detail the datasets used for selecting the few-shot examples and evaluation of the larger VLMs on \ps{} framework. $\mathcal{A}_\mathcal{C}$: caption agent, $\mathcal{A}_\mathcal{E}$: explanation agent, $\mathcal{A}_\mathcal{I}$: intervention agent, $\mathcal{T}_\mathcal{R}$: support set for few-shot exemplars, $\mathcal{T}_\mathcal{S}$: evaluation set.}
\renewcommand{\arraystretch}{1.2}
\setlength{\tabcolsep}{0.9mm}
\begin{tabular}{c|c|c|c}
\hline
\textbf{dataset} & \textbf{task focus} & \textbf{split / label dist.} & \textbf{total} \\ \hline
\multicolumn{4}{c}{{\textbf{phase 1: training task-specific agents ($\mathcal{A}_\mathcal{C}$, $\mathcal{A}_\mathcal{E}$, and $\mathcal{A}_\mathcal{I}$)}}} \\ \hline
\sysMemeCap{} & caption & train + val & 5,828 \\
\sysHateRedAug{} & explanation & H: 2,982 \quad NH: 5,388 & 8,370 \\
\sysMemeSense{} & common sense & train only & 299 \\
\sysMemeSense{} & intervention & train only & 299 \\ \hline \hline
\multicolumn{4}{c}{{\textbf{phase 2: support set ($\mathcal{T}_\mathcal{R}$)}}} \\ \hline
FHM & few-shot examples & same as \sysHateRedAug{} & 8370 \\
MAMI & few-shot examples & H:5000 \quad NH:5000 & 10000 \\ \hline \hline
\multicolumn{4}{c}{{\textbf{phase 2: evaluation  ($\mathcal{T}_\mathcal{S}$)}}} \\ \hline
FHM-extended & C+E+I & H: 485 \quad NH: 498 & 983 \\
MAMI-extended & C+E+I & H: 478 \quad NH: 467 & 945 \\ \hline
\multicolumn{4}{l}{\tiny{*H: hateful/misogynistic, NH: non-hateful/non-misogynistic}}
\end{tabular}
\label{tab:experiment_datasets}
\end{table}

\section{Annotation of datasets}
\label{sec:annotation_datasets}

\noindent\textbf{Dataset for training the agents}: In order to train the explanation agent we use the \sysHateRed{} dataset. However, this set has an explanation for only the hateful memes. We therefore annotate the non-hateful memes with their explanations. For training the caption and intervention generation agents, we use prior datasets -- \sysMemeCap{} and \sysMemeSense{} respectively.\\
\noindent\textbf{Dataset for evaluation}: There is no dataset in the literature that has the classification label, the explanation and the intervention altogether. Hence we annotate the test split of the FHM and MAMI datasets to include explanation and intervention annotations in addition to the labels (e.g., hateful, misogynistic etc). Details of annotators and the process of annotation carried out at each step are described in the upcoming subsections.

\subsection{Annotators}
\label{sec:annotators}
We only selected researchers who have professional working experience and have prior publications in this domain to ensure high-quality annotations. The annotator pool comprises PhD and masters students along with industry professionals and researchers who have been actively involved in prior datasets curation. This led to a strong selective team of six researchers within the age range of 21-30 years.

\subsection{Annotation process}
\label{sec:annotation_process}
For all annotations, we use the instructions cited in prior works (for explanation~\cite{Hee2023DecodingTU}, and for intervention~\cite{jha2024memeguard, adak2025memesenseadaptiveincontextframework}) and used their annotation guidelines. To maintain high quality, we propose a three-staged annotation: (i) pilot stage, (ii) full annotation, and (iii) evaluation and refinement.\\
\noindent(i) \textit{Pilot stage}: In the first stage, each expert annotator was assigned 25 samples randomly taken from the \sysHateRed{} and \sysMemeSense{} datasets. The pilot annotations were then verified in an anonymous manner where on the basis of evaluation judged by the primary author of this paper, more than 95\% of performed annotations by each annotator in pilot phase was found to be accurate and highly aligning to ground truth results. This strengthens the quality of annotations which experts in the domain bring with them.\\
\noindent(ii) \textit{Full annotation workflow}: Once the pilot phase is over we share (a) all the non-hateful {train} points from the FHM (seed dataset of \sysHateRed{}) for explanation annotation and (b) the test points from FHM and MAMI datasets for explanation and intervention annotation. Each annotator was provided with samples from these datasets along with their metadata and the annotation was carried out using the \texttt{Doccano}\footnote{\url{https://github.com/doccano/doccano}} platform. The instructions for these annotations are directly adopted from the respective source papers.\\
\noindent(iii) \textit{Evaluation and refinement}: As stated beforehand, quality was our top most priority, and to ensure that, we again performed an additional quality control by assigning each sample to two independent researchers. All the annotated samples were re-evaluated for their content and correctness; we found that more than 94\% of the samples were agreed upon to be precise with a high agreement score of Cohen's $\kappa=0.932$. On those samples where both evaluators disagree (nearly 2\% samples), were re-written and finally we had a very robust and high quality dataset for training and evaluation purposes. 


\section{Curated dataset}
\label{sec:curated-dataset}

At the end of the annotation process, we have the following datasets curated (see Table~\ref{tab:experiment_datasets} for summary). \\
\noindent\textit{Training datasets for agents --} We train four distinct agents to facilitate the generation of silver data. To train the captioning agent, we utilize the \sysMemeCap{} dataset~\cite{hwang2023memecap} (train + validation splits: 5,828 samples) for meme-oriented captions. For the explanation agent, we combine the \sysHateRed{} dataset~\cite{Hee2023DecodingTU} (2,982 hateful samples) with our manually annotated set of 5,388 non-hateful samples from FHM, ensuring the agent is label-aware and balanced. We call this new dataset with explanations for both hateful and non-hateful memes as \sysHateRedAug{}. Finally, for the commonsense and intervention agents, we leverage the \sysMemeSense{} dataset~\cite{adak2025memesenseadaptiveincontextframework} (299 training samples), which provides ground-truth causal reasoning and interventions for toxic memes. A detailed example is shown in Figure~\ref{fig:main_teaser}. \\
\noindent \textit{Support set for few-shot examples --} For passing the few-shot examples, we use the train splits of the FHM and MAMI datasets. Since FHM is the seed dataset of \sysHateRedAug{}, we use its samples directly, and for MAMI, we utilize its 10000 training examples out of which 5000 memes are misogynistic and 5000 memes are non-misogynistic. Since this support set is taken verbatim from the existing standard datasets in the literature, no deduplication is additionally required.\\
\noindent\textit{Evaluation datasets --} For the final evaluation of our unified framework (classification + explanation + intervention), we utilize the test splits of two popular benchmark datasets: FHM~\cite{kiela2020hateful} and MAMI~\cite{fersini2022semeval}. As noted in Section~\ref{sec:annotation_datasets}, we extend the ground truth for these test splits to include explanation and intervention annotations. Specifically, we use the FHM test\_seen split (983 total; 485 hateful, 498 non-hateful) and the MAMI test split (945 total; 478 misogynistic, 467 non-misogynistic).


\section{The \ps{} framework}
\label{sec:experiments}
In this section, we outline how we build the task specific agents, and use the silver data obtained from these agents to fuel a few-shot framework. This few-shot model is simultaneously able to perform classification and generation of explanation and intervention for a test sample.

\subsection{Task specific agents}
\label{subsec:agents}
\begin{figure}[t]
    \centering
    \includegraphics[width=1\columnwidth]{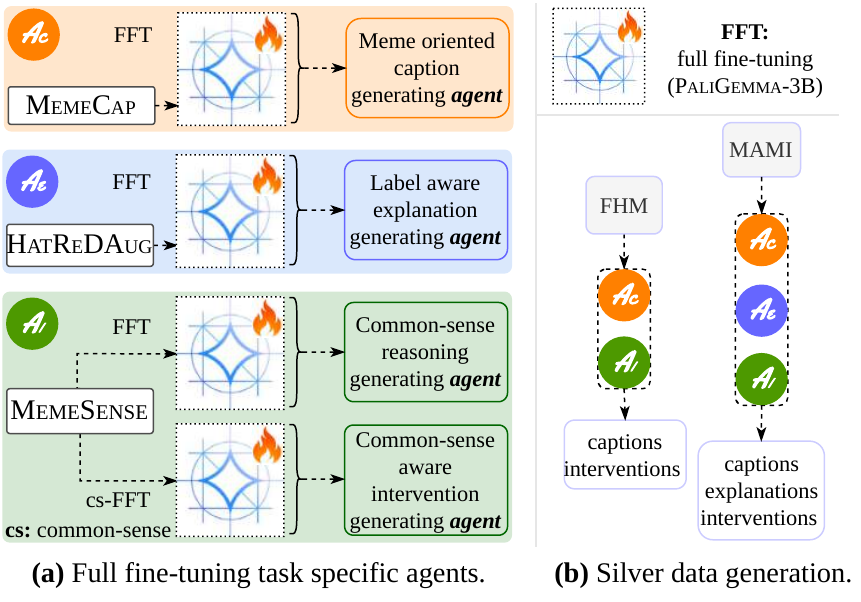}
    \caption{\footnotesize Overview of fine-tuning task-specific agents and using them for silver data generation on the support set $\mathcal{T}_\mathcal{R}$ of FHM and MAMI datasets. Note that in the case of generating explanations, we use agent $\mathcal{A}_\mathcal{E}$ only for MAMI dataset and use \textsc{\sysHateRedAug{}} explanations for FHM.
    }
    \label{fig:silver_data}
\end{figure}

We train three different types of agents as shown in Figure~\ref{fig:silver_data}. We specifically train the \sysP{} 3 billion parameters model so that it can be trained effectively in a low-resource and adaptable setup. These three agentic modules, their respective input and output formulation and the datasets on which they are trained are provided below. For all these small agents, we perform full fine-tuning on the model checkpoint: \texttt{paligemma-3b-pt-448}. Detailed experimental setup is provided in subsection~\ref{subsec:models_and_metrics}.\\
\noindent\textit{Meme oriented caption generation --} Memes significantly depart compared to general images as their contextualized meaning is portrayed as a combination of embedded text and the image itself. Numerous examples have been covered in the \sysMemeCap{} paper~\cite{hwang2023memecap}, where they have shown that the general description of a meme failed to uncover the actual meaning being conveyed. Therefore, we fine-tune an agent to provide meme-oriented captions for a given input. To this purpose, we use the dataset \sysMemeCap{} to train our model and input the meme to generate the captions as output during full fine-tuning. We call this trained agent $\mathcal{A}_\mathcal{C}$.\\
\noindent\textit{Label aware explainability generator --} We fine-tune a model to enact as a label-aware explainability generator to generate an explanation given the ground truth label of the meme. We use \sysHateRedAug{} to fine-tune the agent.
The reason that we train a label-aware explanation generator is that we want to generate very good silver training data for the train split of unseen datasets, which will be used for the few-shot setup later. We call this trained agent $\mathcal{A}_\mathcal{E}$.\\
\noindent\textit{Intervention generator for hateful memes --} As introduced in the \sysMemeSense{} paper~\cite{adak2025memesenseadaptiveincontextframework}, a set of common sense reasoning aligned to human thought process can appropriately express why a given sample is classified as hateful. We use this dataset to first fine-tune a common sense reasoning generator and then use these to further generate the actual interventions. Here also we use full fine-tuning for both common sense reasoning and intervention generation.  We call this trained agent $\mathcal{A}_\mathcal{I}$.

\subsection{Agents guided few-shot framework}
\label{subsec:few_shot}
The few-shot prompting has two major steps -- (a) exemplar selection, and (b) enrichment of the selected exemplars. We describe each of these steps below.\\
\noindent\textit{Exemplar selection}: Let the test sample be denoted as $t$. Let the few-shot examples corresponding to $t$ be denoted by the set $\mathcal{F} = \{f_i\}_{i=1}^{n}$. The few-shot examples $f_i$ are chosen such that they have the highest cosine similarity with the $t$ in terms of the \sysSL{} embeddings\footnote{We also present results assuming \textsc{CLIP} and \textsc{BLIP} embeddings in Appendix~\ref{app:embed_retrievers}.}. Note that in Table~\ref{tab:experiment_datasets}, $\mathcal{T}_\mathcal{R}$ is the support set from which few-shot examples are selected, and $\mathcal{T}_\mathcal{S}$ is the test set from which a query sample $t$ is selected. We present results for $n=\{2, 4, 8\}$. \\
\noindent\textit{Enrichment of the selected exemplars}: For each selected exemplar, $f_i$, we pass it through the different agents introduced earlier. Thus $\mathcal{A}_\mathcal{C}$($f_i$) returns the caption of the exemplar $f_i$. Similarly, $\mathcal{A}_\mathcal{E}$($f_i$) returns the explanation for $f_i$ and $\mathcal{A}_\mathcal{I}$($f_i$) returns the intervention only if $f_i$ is a hateful meme. Each $f_i$ is enriched with these additional (silver) information to enable multi-tasking and also improve the prediction performance. Finally, for the test sample $t$ we predict the label, the explanation and the intervention (if $t$ is predicted as hateful).

\noindent Further, we have taken care of these aspects while running our experiments: All the agents ($\mathcal{A}_\mathcal{C}$, $\mathcal{A}_\mathcal{E}$, and $\mathcal{A}_\mathcal{I}$) for the explanation, caption, and intervention generation are on the training data points $\mathcal{T}_\mathcal{R}$. For the test points $\mathcal{T}_\mathcal{S}$, no explanation or intervention has been generated. Only the caption has been generated. In the few-shot setup, the few-shot examples sampled from $\mathcal{T}_\mathcal{R}$ are fed along with explanation, intervention, caption and ground truth label. The test point sampled from $\mathcal{T}_\mathcal{S}$ is passed without any of these except the caption, and thus, there is no leakage during few-shot prompting. Hence, label overlap for the test sample is not possible in these cases. Further, as mentioned earlier, for FHM $\mathcal{T}_\mathcal{R}$ we make use of the explanations from \sysHateRedAug{} (as it has the same seed dataset - FHM); the explanation agent $\mathcal{A}_\mathcal{E}$ is only used for the MAMI $\mathcal{T}_\mathcal{R}$ since the training samples do not have ground truth explanations.

\noindent \textbf{Low resource agentic steering:} We use the term agentic steering because each module ($\mathcal{A}_\mathcal{C}$, $\mathcal{A}_\mathcal{E}$, and $\mathcal{A}_\mathcal{I}$) operates autonomously with a well-defined objective, processes its own inputs, generates task-specific outputs, and influences the behavior of a downstream system. The agents exhibit a clear division of responsibilities: caption generation, explanation generation, and intervention generation. Furthermore, the intervention module performs multi-stage reasoning by leveraging commonsense knowledge before generating interventions, reflecting a structured decision-making process beyond a single forward pass. Most importantly, these agents do not modify the parameters of the target black-box VLM. They influence its behavior externally through the information they contribute to the few-shot context. In this sense, they act as autonomous components that perceive, reason, and generate outputs that guide the downstream model toward improved classification, explanation, and intervention. Importantly, only the captioning agent ($\mathcal{A}_\mathcal{C}$) is executed at inference time; all explanations and interventions for few-shot exemplars are pre-computed and cached offline. Thus, deployment requires only a single 3B-parameter model alongside a frozen closed-source model like \sysGPT{} with >200B parameters, enabling a model roughly \textbf{67× smaller} to effectively steer a much larger black-box VLM. The exact prompt for generating the responses using the \ps{} framework and the corresponding definitions of FHM/MAMI datasets are noted in Appendix~\ref{app:prompts}.

\subsection{Models and evaluation metrics}
\label{subsec:models_and_metrics}
\noindent\textit{Deployed models --} As noted earlier, for the small agents we perform full fine-tuning of \sysP{} model with 3B parameters.  We employed \texttt{learning\_rate} of 2e-5 with Adam optimizer and a \texttt{weight\_decay} of 1e-6 for 2 epochs. At both stages (fine-tune and inference) and for all task specific agents, we use the standard bfloat-16 representation for weights and bias representation. For the few-shot prompting, we use two open source models \sysI{} (8B parameters, checkpoint: \texttt{OpenGVLab/InternVL3\_5-8B-HF}), \sysPx{} (12B parameters, \texttt{mistralai/Pixtral-12B-2409}) and the proprietary \sysGPT{} model. For all inference experiments, we run the models on a very low temperature value of 0.001 to ensure reproducibility. The \texttt{max\_new\_tokens} parameter is tuned to 100 tokens. For \sysGPT{}, we use the Microsoft Azure services and deploy the open-source models \sysI{}, \sysPx{}, and \sysP{} on our server with bfloat-16 representation.

\noindent\textit{Evaluation metrics --} For classification, we relied on accuracy and macro-F1 score, as done in previous works as well~\cite{Rizwan_Bhaskar_Das_Majhi_Saha_Mukherjee_2025}. For explanation and intervention, we use \sysRG{}-L~\cite{lin2004rouge} score, \sysCSIM{}, and \sysSBERT{}-F1~\cite{zhang2020bertscoreevaluatingtextgeneration} as carried out in previous tasks as well. For \sysCSIM{} we utilize the \texttt{all-MiniLM-L6-v2} checkpoint from \textsc{Sentence-BERT}\footnote{\url{https://sbert.net/}}. In the next section, we present detailed results, noting multiple takeaways.

\subsection{Baselines}

We benchmark our results on an array of prior works. For classification, multiple frameworks are compared: \textsc{PromptHate}~\cite{cao2023promptingmultimodalhatefulmeme}, \textsc{ProCap}~\cite{cao2023pro}, \textsc{ModHate}~\cite{cao2024modularizednetworksfewshothateful}, \textsc{Few-Shot}~\cite{hee2024bridging}, U-CoT+~\cite{pan2026readseeguidingunimodal}, M2KE~\cite{lu2025having}, \textsc{Vlm-Lim}~\cite{Rizwan_Bhaskar_Das_Majhi_Saha_Mukherjee_2025}, and \textsc{LoReHM}~\sysGPT{}~\cite{huang2024towards}. While \textsc{PromptHate} uses prompting, \textsc{Pro-Cap} improves upon vision-language models for better classification. However, they require rigorous fine-tuning for the alignment of image and text modalities. \textsc{Few-Shot}, \textsc{ModHate}, and \textsc{LoReHM} present low-resource setups anchored upon few-shot and in-context learning; \textsc{Vlm-Lim} presents a complete zero-shot setup on similar lines and U-CoT+ presents a reasoning-aligned classification framework using two models, a VLM and an LLM. \textsc{M2KE} is one of the most recent works that utilizes a two-staged framework with adaptive knowledge fusion for obtaining better results; however, it requires significant fine-tuning. For explanation, we consider the framework discussed in \sysHateRed{}~\cite{Hee2023DecodingTU} and to ensure fair assessment, we fine-tune this framework using \sysHateRedAug{}. For intervention, we utilize a recent work \sysMemeSense{}~\cite{adak2025memesenseadaptiveincontextframework} that proposes a low-resource setup for intervention generation.


\section{Results}
\label{sec:results}


\label{subsec:evaluate_agents}

\begin{table}[t]
\centering
\caption{\footnotesize Performance of the four task-specific fine-tuned \sysP{}-3B agents. Higher values indicate strong meaning preservation despite lexical variation. Here, cap: captions, exp: explanation, c-s: common sense, int: intervention, rgL: \sysRG{}-L, ss: \sysCSIM{}, and bsf1: \sysSBERT{}-F1.}
\scriptsize
\renewcommand{\arraystretch}{1.20}
\setlength{\tabcolsep}{2mm}

\resizebox{\linewidth}{!}{
\begin{tabular}{c|c|c|c|c|c}
\hline
\rowcolor[HTML]{DAE8FC}
\textbf{agent} & \textbf{dataset} & \textbf{\# samples} & \textbf{rgL} & \textbf{ss} & \textbf{bsf1} \\ \hline

cap      & \sysMemeCap{}       & 559 & 0.3571 & 0.6673 & 0.9042 \\
exp     & \sysHateRed{}       & 246 & 0.3777 & 0.6326 & 0.9079 \\
c-s    & \sysMemeSense{}     & 134 & 0.2501 & 0.7212 & 0.8999 \\
int    & \sysMemeSense{}     & 134 & 0.3438 & 0.7857 & 0.9114 \\
\hline
\end{tabular}
}
\label{tab:agent_eval_app_color}
\end{table}

\noindent\textbf{Robustness of task-specific agents}: Here we present an evaluation of the four task-specific \sysP{}-3B agents introduced in Section~\ref{subsec:agents} (also refer to Figure~\ref{fig:silver_data}). These agents are responsible for (i) $\mathcal{A}_\mathcal{C}$-- meme-oriented captioning, (ii) $\mathcal{A}_\mathcal{E}$-- label-aware explanation, (iii) $\mathcal{A}_\mathcal{I}$-- common-sense reasoning and intervention generation. We evaluate each agent on its held-out test split ($n=559$ for \sysMemeCap{}, $n=246$ for \sysHateRed{}, and $n=134$ for \sysMemeSense{}) using the three metrics that jointly capture surface-level and semantic fidelity -- \sysRG{}, \sysCSIM{}, and \sysSBERT{}-F1. We obtain the following key insights from the results presented in Table~\ref{tab:agent_eval_app_color}: (i) \textit{Robust semantic alignment across all agents}: \sysSBERT{}-F1 values around 0.90 indicate that the generated content closely preserves meaning even when surface form varies. (ii) \textit{Commonsense reasoning shows expected lexical diversity}: The \sysMemeSense{} agent yields the lowest \sysRG{}-L, which is consistent with the open-ended nature of commonsense rationales, mirroring observations from the \sysMemeSense{} framework. (iii) \textit{Interventions are structurally stable and semantically precise}: The intervention agent exhibits the highest semantic coherence, reflecting the structured and safety-centered nature of intervention writing.\\
\noindent Conclusively, these task-specific agents are fairly robust and can enact as strong guides for the larger VLMs.

\begin{table*}[ht]
\footnotesize
\centering
\caption{\footnotesize Results comparing our method with the previous works. `-' here represent that the prior works cannot perform those mentioned tasks. C: Classification, E: Explanation, I: Intervention, PH: \textsc{PromptHate}, PC: \textsc{Pro-Cap}, MH: \textsc{Mod-Hate}, FS: \textsc{Few-Shot}, UC: \textsc{U-CoT+}, MK: \textsc{M2KE}, V-L: \textsc{Vlm-Lim} (\sysGPT{}), LM: \textsc{LoReHM} (\sysGPT{}), HR: \sysHateRed{}, MS: \sysMemeSense{}, IVL: \sysI{}, PX: \sysPx{}, GPT: \sysGPT{}, acc: accuracy, rgL: \sysRG{}-L, ss: \sysCSIM{}, bsf1: \sysSBERT{}-F1, sprt: support. Best results are marked \colorbox[HTML]{D9EAD3}{\textbf{green}} and second best results are marked \colorbox[HTML]{F1FCED}{\ul light green}.}
\setlength{\tabcolsep}{0.39mm}
\renewcommand{\arraystretch}{1.2}
\begin{tabular}{c|cc|cc|ccc|cccc|cc|ccc|cccc}
\rowcolor[HTML]{FFF2CC} 
\cellcolor[HTML]{D9D2E9}                                                                                                           & \multicolumn{2}{c|}{\cellcolor[HTML]{D9D2E9}}                                                  & \multicolumn{2}{c|}{\cellcolor[HTML]{FFF2CC}\textbf{FHM-C}}                   & \multicolumn{3}{c|}{\cellcolor[HTML]{FFF2CC}\textbf{FHM-E}}                                                            & \multicolumn{4}{c|}{\cellcolor[HTML]{FFF2CC}\textbf{FHM-I}}                                                                                                         & \multicolumn{2}{c|}{\cellcolor[HTML]{FFF2CC}\textbf{MAMI-C}}                  & \multicolumn{3}{c|}{\cellcolor[HTML]{FFF2CC}\textbf{MAMI-E}}                                                           & \multicolumn{4}{c}{\cellcolor[HTML]{FFF2CC}\textbf{MAMI-I}}                                                                                                  \\ \cline{4-21} 
\rowcolor[HTML]{FCE5CD} 
\multirow{-2}{*}{\cellcolor[HTML]{D9D2E9}\textbf{task}}                                                                            & \multicolumn{2}{c|}{\multirow{-2}{*}{\cellcolor[HTML]{D9D2E9}\textbf{model}}}                  & \textbf{acc}                           & \textbf{mf1}                           & \textbf{rgL}                           & \textbf{ss}                            & \textbf{bsf1}                          & \textbf{rgL}                           & \textbf{ss}                            & \textbf{bsf1}                          & \textbf{sprt}                              & \textbf{acc}                           & \textbf{mf1}                           & \textbf{rgL}                           & \textbf{ss}                            & \textbf{bsf1}                          & \textbf{rgL}                          & \textbf{ss}                            & \textbf{bsf1}                          & \textbf{sprt}                        \\ \hline
                                                                                                                                   & \multicolumn{2}{c|}{PH}                                                       & 72.98                                  & 72.24                                  & -                                      & -                                      & -                                      & -                                      & -                                      & -                                      & -                                          & 70.31                                  & 70.18                                  & -                                      & -                                      & -                                      & -                                     & -                                      & -                                      & -                                    \\
                                                                                                                                   & \multicolumn{2}{c|}{PC}                                                          & 75.1                                   & 74.85                                  & -                                      & -                                      & -                                      & -                                      & -                                      & -                                      & -                                          & 73.63                                  & 73.42                                  & -                                      & -                                      & -                                      & -                                     & -                                      & -                                      & -                                    \\
                                                                                                                                   & \multicolumn{2}{c|}{MH}                                                         & 57.6                                   & 53.88                                  & -                                      & -                                      & -                                      & -                                      & -                                      & -                                      & -                                          & 69.05                                  & 68.78                                  & -                                      & -                                      & -                                      & -                                     & -                                      & -                                      & -                                    \\
                                                                                                                                   & \multicolumn{2}{c|}{FS}                                               & 66                                     & 65.8                                   & -                                      & -                                      & -                                      & -                                      & -                                      & -                                      & -                                          & 70.5                                   & 70.1                                   & -                                      & -                                      & -                                      & -                                     & -                                      & -                                      & -            
                                                                                                                            \\
                                                                                                                                   & \multicolumn{2}{c|}{UC}                                               & 74.06                                     & 74.05                                   & -                                      & -                                      & -                                      & -                                      & -                                      & -                                      & -                                          & 76.72                                   & 76.34                                   & -                                      & -                                      & -                                      & -                                     & -                                      & -                                      & -   \\
                                                                                                                                   & \multicolumn{2}{c|}{MK}                                                             & 75.76                                  & 75.62                                  & -                                      & -                                      & -                                      & -                                      & -                                      & -                                      & -                                          & 75.85                                  & 75.71                                  & -                                      & -                                      & -                                      & -                                     & -                                      & -                                      & -                                    \\
                                                                                                                                   & \multicolumn{2}{c|}{V-L}                                              & 73.4                                   & 73.1                                   & -                                      & -                                      & -                                      & -                                      & -                                      & -                                      & -                                          & 83.6                                   & 83.54                                  & -                                      & -                                      & -                                      & -                                     & -                                      & -                                      & -                                    \\
\multirow{-7}{*}{\textbf{C}}                                                                                          & \multicolumn{2}{c|}{LM}                                                  & 70.2                                   & 70.14                                  & -                                      & -                                      & -                                      & -                                      & -                                      & -                                      & -                                          & 83                                     & 82.98                                  & -                                      & -                                      & -                                      & -                                     & -                                      & -                                      & -                                    \\ \hline
\textbf{E}                                                                                                               & \multicolumn{2}{c|}{HR}                                                          & -                                      & -                                      & 0.126                                    & 0.417                                    & 0.862                                    & -                                      & -                                      & -                                      & -                                          & -                                      & -                                      & 0.105                                    & 0.361                                    & 0.859                                    & -                                     & -                                      & -                                      & -                                    \\ \hline
\textbf{I}                                                                                                               & \multicolumn{2}{c|}{MS}                                                          & -                                      & -                                      & -                                    & -                                    & -                                    & 0.204                                      & 0.689                                      & 0.88                                      & -                                          & -                                      & -                                      & -                                    & -                                    & -                                    & 0.225                                     & 0.73                                      & 0.886                                      & -                                    \\ \hline \hline
                                                                                                             & \multicolumn{1}{c|}{}                                     & \cellcolor[HTML]{D9D2E9}\textbf{0} & 65.21                                  & 65.1                                  & 0.183                                  & 0.508                                  & 0.878                                   & 0.103                                  & 0.323                                  & 0.86                                  & 348                                        & 69.42                                  & 67.53                                     & 0.193                                  & 0.513                                   & 0.883                                  & 0.104                                 & 0.298                                  & 0.857                                   & 442                                  \\                      & \multicolumn{1}{c|}{}                                     & \cellcolor[HTML]{D9D2E9}\textbf{2} & 71.41                                  & 71.41                                  & 0.209                                  & 0.599                                  & 0.88                                   & 0.215                                  & 0.58                                  & 0.888                                  & 341                                        & 71.96                                  & 71.69                                     & 0.207                                  & 0.581                                   & 0.885                                  & 0.332                                 & 0.792                                  & 0.915                                   & 386                                  \\
                                                                                                                                   & \multicolumn{1}{c|}{}                                     & \cellcolor[HTML]{D9D2E9}\textbf{4} & 73.55                                  & 73.43                                  & 0.215                                  & 0.625                                  & 0.883                                  & 0.263                                  & 0.701                                  & 0.901                                  & 324                                        & 74.39                                  & 74.24                                  & 0.186                                  & 0.569                                  & 0.882                                  & 0.373                                 & 0.83                                  & 0.918                                  & 388                                  \\
                                                                                                                                   & \multicolumn{1}{c|}{\multirow{-3}{*}{IV}} & \cellcolor[HTML]{D9D2E9}\textbf{8} & 71.41                                  & 70.33                                  & 0.227                                  & 0.635                                  & 0.887                                  & \cellcolor[HTML]{D9EAD3}{ \textbf{0.29}}    & \cellcolor[HTML]{D9EAD3}{ \textbf{0.777}}    & \cellcolor[HTML]{D9EAD3}{ \textbf{0.907}}    & 252                                        & 75.34                                  & 75.25                                  & 0.163                                  & 0.557                                  & 0.879                                  & \cellcolor[HTML]{D9EAD3}\textbf{0.395} & \cellcolor[HTML]{D9EAD3}{\ul 0.843} & \cellcolor[HTML]{D9EAD3}{\ul 0.919} & 385                                  \\ \cline{2-21} 
                                                                                                                             & \multicolumn{1}{c|}{}                                     & \cellcolor[HTML]{D9D2E9}\textbf{0} & 67.85                                  & 67.83                                  & 0.158                                  & 0.487                                   & 0.868                                   & 0.136                                  & 0.494                                  & 0.87                                  & 345                                        & 74.07                                  & 73.31                                  & 0.17                                  & 0.462                                  & 0.872                                  & 0.138                                 & 0.53                                  & 0.869                                  & 429                                  \\
                                                                                                                             & \multicolumn{1}{c|}{}                                     & \cellcolor[HTML]{D9D2E9}\textbf{2} & 69.79                                  & 69.78                                  & 0.197                                  & 0.56                                   & 0.88                                   & 0.213                                  & 0.611                                  & 0.887                                  & 338                                        & 67.94                                  & 65.88                                  & 0.191                                  & 0.513                                  & 0.882                                  & 0.326                                 & 0.802                                  & 0.912                                  & 437                                  \\

                                                                                                                                   & \multicolumn{1}{c|}{}                                     & \cellcolor[HTML]{D9D2E9}\textbf{4} & 70.3                                   & 70.26                                  & 0.211                                  & 0.606                                  & 0.884                                  & 0.249                                  & 0.699                                  & 0.897                                  & 329                                        & 73.65                                     & 72.89                                  & 0.19                                   & 0.536                                  & 0.881                                  & 0.358                                 & 0.833                                  & 0.918                                  & 427                                  \\
                                                                                                                                   & \multicolumn{1}{c|}{\multirow{-3}{*}{PX}}   & \cellcolor[HTML]{D9D2E9}\textbf{8} & 72.53                                  & 72.45                                  & 0.224                                  & 0.635                                  & 0.887                                  & \cellcolor[HTML]{F1FCED}{\ul 0.272} & \cellcolor[HTML]{F1FCED}{\ul 0.774} & \cellcolor[HTML]{F1FCED}{\ul 0.903} & 330                                        & 75.42                                  & 74.99                                  & 0.17                                  & 0.55                                  & 0.879                                  & \cellcolor[HTML]{F1FCED}{\ul 0.391}   & \cellcolor[HTML]{F1FCED}\textbf{ 0.849}     & \cellcolor[HTML]{F1FCED}\textbf{ 0.921}    & 418                                  \\ \cline{2-21} 
                                                                                                                             
                                                                                                                       & \multicolumn{1}{c|}{}                                     & \cellcolor[HTML]{D9D2E9}\textbf{0} & 75.89                                  & 75.89                                  & 0.195                                  & 0.549                                  & 0.879                                  & 0.126                                  & 0.418                                  & 0.866                                  & 376                                        & 86.35                                  & 86.33                                  & 0.224                                  & 0.539                                  & 0.882                                  & 0.127                                 & 0.47                                  & 0.865                                  & 424                                  \\            & \multicolumn{1}{c|}{}                                     & \cellcolor[HTML]{D9D2E9}\textbf{2} & 79.35                                  & 79.34                                  & 0.221                                  & 0.648                                  & 0.886                                  & 0.144                                  & 0.465                                  & 0.871                                  & 400                                        & 88.25                                  & 88.23                                  & 0.224                                  & 0.615                                  & 0.886                                  & 0.139                                 & 0.508                                  & 0.869                                  & 438                                  \\
                                                                                                                                   & \multicolumn{1}{c|}{}                                     & \cellcolor[HTML]{D9D2E9}\textbf{4} & \cellcolor[HTML]{F1FCED}{\ul 79.45}    & \cellcolor[HTML]{F1FCED}{\ul 79.44}    & \cellcolor[HTML]{F1FCED}{\ul 0.233}    & \cellcolor[HTML]{F1FCED}{\ul 0.674}    & \cellcolor[HTML]{F1FCED}{\ul 0.89}     & 0.155                                  & 0.491                                  & 0.875                                  & \cellcolor[HTML]{F1FCED}{\ul 401}          & \cellcolor[HTML]{F1FCED}{\ul 88.78}    & \cellcolor[HTML]{F1FCED}{\ul 88.75}     & \cellcolor[HTML]{F1FCED}{\ul 0.231}    & \cellcolor[HTML]{F1FCED}{\ul 0.641}    & \cellcolor[HTML]{F1FCED}{\ul 0.888}    & 0.172                                 & 0.565                                  & 0.877                                  & \cellcolor[HTML]{F1FCED}{\ul 444}    \\
\multirow{-9}{*}{\textbf{\begin{tabular}[c]{@{}c@{}}C\\ +\\ E\\ +\\ I\\\end{tabular}}} & \multicolumn{1}{c|}{\multirow{-3}{*}{GPT}}    & \cellcolor[HTML]{D9D2E9}\textbf{8} & \cellcolor[HTML]{D9EAD3}\textbf{80.26} & \cellcolor[HTML]{D9EAD3}\textbf{80.25} & \cellcolor[HTML]{D9EAD3}\textbf{0.242} & \cellcolor[HTML]{D9EAD3}\textbf{0.679} & \cellcolor[HTML]{D9EAD3}\textbf{0.891} & 0.195                                  & 0.563                                  & 0.883                                  & \cellcolor[HTML]{D9EAD3}{\textbf{409}} & \cellcolor[HTML]{D9EAD3}\textbf{89.1} & \cellcolor[HTML]{D9EAD3}\textbf{89.07} & \cellcolor[HTML]{D9EAD3}\textbf{0.238} & \cellcolor[HTML]{D9EAD3}\textbf{0.654} & \cellcolor[HTML]{D9EAD3}\textbf{0.891} & 0.239                                  & 0.693                                  & 0.894                                  & \cellcolor[HTML]{D9EAD3}\textbf{446}\\
\hline
\end{tabular}
\label{tab:results}
\end{table*}

\noindent\textbf{Evaluation on the overall \ps{} framework}: In this subsection, we present the results obtained from our few-shot framework considering \sysSL{} embeddings to measure the similarity between $t$ and $f_i$ to select the set of best exemplars ($\mathcal{F}$). The results are summarized in Table~\ref{tab:results}. The enriched few-shot samples + \sysGPT{} model outperforms all the previous baselines in terms of classification accuracy. These include \textsc{Vlm-Lim} and \textsc{LoReHM}, which are also \sysGPT{} based. Further, this setup also surpasses the fine-tuning based approaches like \textsc{PromptHate}, \textsc{Pro-cap}, and \textsc{M2KE} by a margin of nearly 8\%, 5.4\%, and 4.6\% on the FHM dataset and by a margin of nearly 19\%, 15.5\%, and 12.5\% on the MAMI dataset, respectively and bypasses U-CoT+ -- that uses two models -- by a margin of nearly 6\% on FHM and 13\% on MAMI dataset (see Appendix~\ref{app:significance} for statistical significance tests). On the low-resource setups not using \sysGPT{} like \textsc{Mod-Hate}, and \textsc{Few-Shot}, even the two open-source models \sysI{} and \sysPx{} report improved results by a substantial margin. \sysGPT{} provides significantly better explanations compared to other open source models and the \sysHateRed{} framework. For intervention generation, however, \sysI{} and \sysPx{} turn out to be the best models, outperforming \sysGPT{} and \sysMemeSense{}. Finally, the zero-shot performance of the employed VLMs (where we do not have the benefits of the three helper agents) is much worse compared to their 8-shot counterparts. This further demonstrates the importance of the assembly of help from these agents constituting the \ps{} framework in boosting the overall performance. In the next section, we further discuss various interesting properties of the explanation and intervention texts.\\
\noindent\textbf{Human evaluation}: In this subsection, we present expert evaluations of the quality of the generated explanations and interventions from \ps{}. We randomly sample 300 memes, out of which 200 are for evaluating explanations and 100 are for evaluating interventions. We then provide these samples to two different expert annotators and ask them to rate the quality of the generations on a scale of 10. Annotators were specifically asked to follow the definition of semantic similarity\footnote{\url{https://medium.com/analytics-vidhya/semantic-similarity-in-sentences-and-bert-e8d34f5a4677}} as per the standard documentation. Since \sysI{} and \sysPx{} have followed almost the same pattern, we perform human evaluation only on \sysI{}. Following are the mean semantic similarity scores in the format (\texttt{expert1}, \texttt{expert2}) we obtained for explanation (normalized between [$0,1$]): \sysGPT{} on FHM -- (0.708, 0.72), on MAMI (0.736, 0.756), and for \sysI{} on FHM-- (0.654, 0.654), on MAMI (0.672, 0.688). Similarly, the following are the scores obtained for intervention:  for \sysGPT{} on FHM -- (0.568, 0.592), on MAMI (0.624, 0.608), and for \sysI{} on FHM -- (0.672, 0.644), on MAMI (0.748, 0.736). The results show that the explanations and interventions generated from \ps{} framework aligns well with human judgments. Further, as observed in Table~\ref{tab:results}, \sysGPT{} generates better explanations while open-source models like \sysI{} generate better interventions.\\
\begin{table}[t]
\scriptsize
\centering
\caption{\footnotesize Mean (standard deviation) toxicity of generated explanation and intervention using \textsc{PerspectiveAPI}. exp: explanation and int: intervention.}
\setlength{\tabcolsep}{2mm}
\renewcommand{\arraystretch}{1.35}
\begin{tabular}{c|c|c|c}
\hline
\rowcolor[HTML]{DAE8FC}
\textbf{type}                                   & \textbf{model}             & \textbf{dataset} & \textbf{toxicity score} \\ \hline
\multirow{6}{*}{\textbf{exp}}  & \multirow{2}{*}{\sysGPT{}}       & FHM     & 0.17 (0.15)          \\
                                       &                            & MAMI    & 0.18 (\textit{0.14})         \\ \cline{2-4} 
                                       & \multirow{2}{*}{\sysI{}} & FHM     & 0.15 \textit{(0.17)}          \\
                                       &                            & MAMI    & 0.27 (\textit{0.22})         \\ \cline{2-4} 
                                       & \multirow{2}{*}{\sysPx{}}   & FHM     & 0.2 (\textit{0.2})           \\
                                       &                            & MAMI    & 0.29 (\textit{0.21})         \\ \hline
\multirow{6}{*}{\textbf{int}} & \multirow{2}{*}{\sysGPT{}}       & FHM     & 0.07 (0.05)          \\
                                       &                            & MAMI    & 0.09 (\textit{0.06})          \\ \cline{2-4} 
                                       & \multirow{2}{*}{\sysI{}} & FHM     & 0.11 (\textit{0.06})          \\
                                       &                            & MAMI    & 0.15 (\textit{0.07})          \\ \cline{2-4} 
                                       & \multirow{2}{*}{\sysPx{}}   & FHM     & 0.14 (\textit{0.07})          \\
                                       &                            & MAMI    & 0.16 (\textit{0.07}) \\ \hline        
\end{tabular}
\label{tab:toxicity}
\end{table}
\noindent\textbf{Toxicity evaluation}: The generated explanations and interventions should be themselves safe and non-toxic.  We use \textsc{PerspectiveAPI}\footnote{\url{https://developers.perspectiveapi.com/s/about-the-api-score}} to test if this is true. We use its benchmark \texttt{TOXICITY} attribute for toxicity estimation (range [$0,1$], score $\ge$ 0.7 means toxic, score $\le$ 0.3 means non-toxic)~\cite{10.1145/3746027.3754553}. In Table~\ref{tab:toxicity}, we present the mean toxicity score along with the standard deviation for all models in the 8-shot setup. None of the mean toxicity scores exceeds 0.3, which confirms that the toxicity level for the generated explanation and intervention remains low and is suitable for \textit{real-time} deployment purposes. The intervention texts are the least toxic, which is very necessary as they are the directly user-facing.


\section{Analysis}
\label{sec:analysis}
\begin{figure}[t]
    \centering
    \includegraphics[width=1\linewidth]{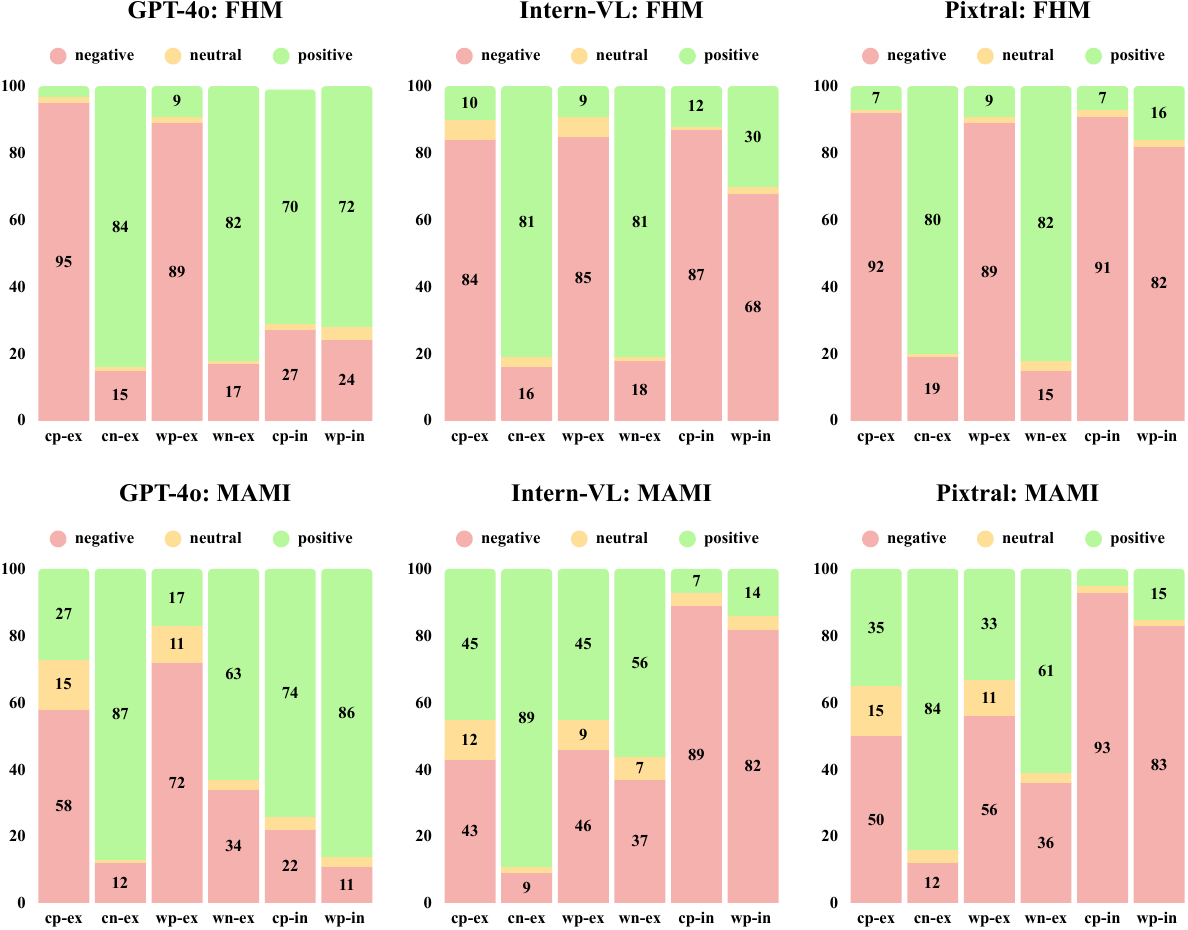}
    \caption{\footnotesize Bar charts presenting the distribution of sentiments for all models and across both datasets. Here, \texttt{cp}: correct positive, \texttt{cn}: correct negative, \texttt{wp}: wrong positive, \texttt{wn}: wrong negative, \texttt{ex}: explanation, \texttt{in}: intervention.}
    \Description{Sentiment analysis over different prediction types.}
    \label{fig:sentiment_analysis}
\end{figure}

\begin{table*}[!ht]
\centering
\setlength{\tabcolsep}{0.33mm}
\begin{tabular}{ccc}
\textbf{\includegraphics[width=0.33\linewidth]{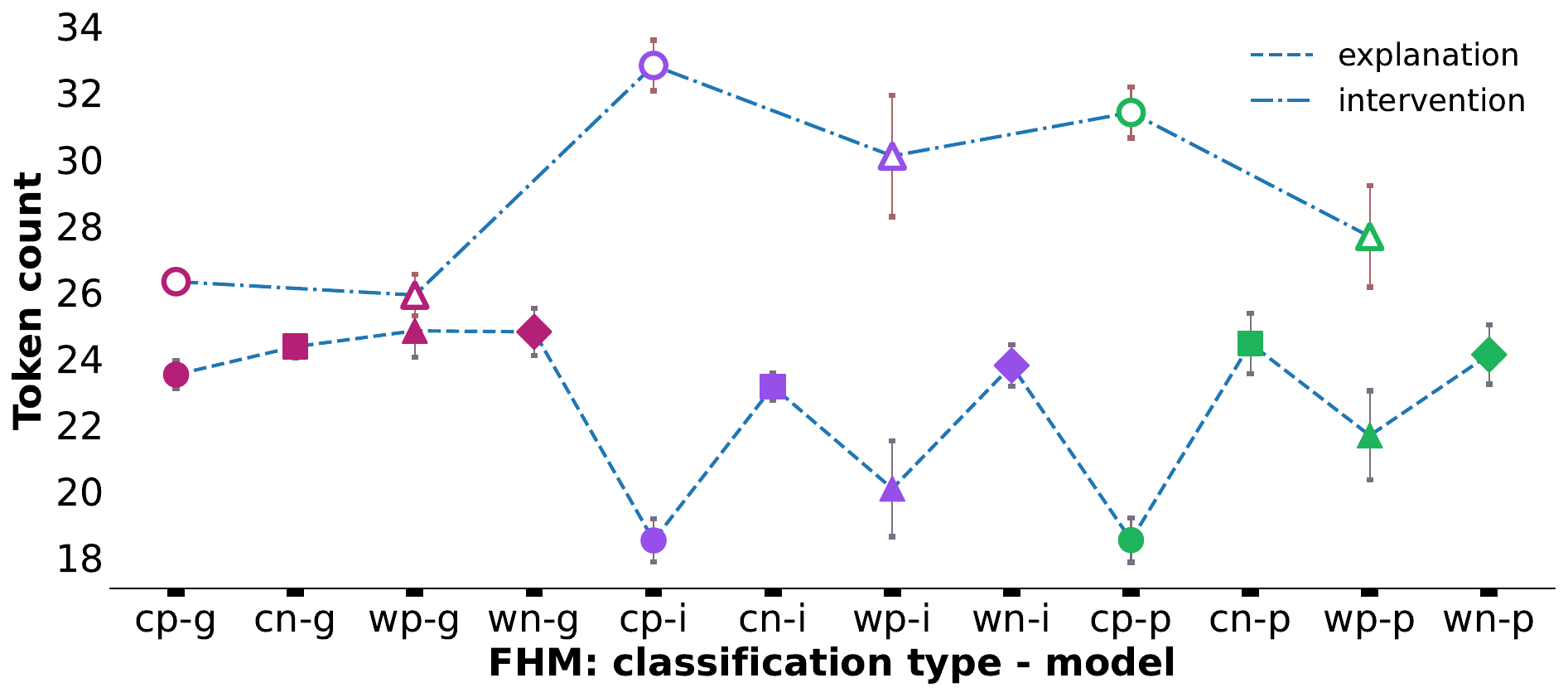}} &
\textbf{\includegraphics[width=0.33\linewidth]{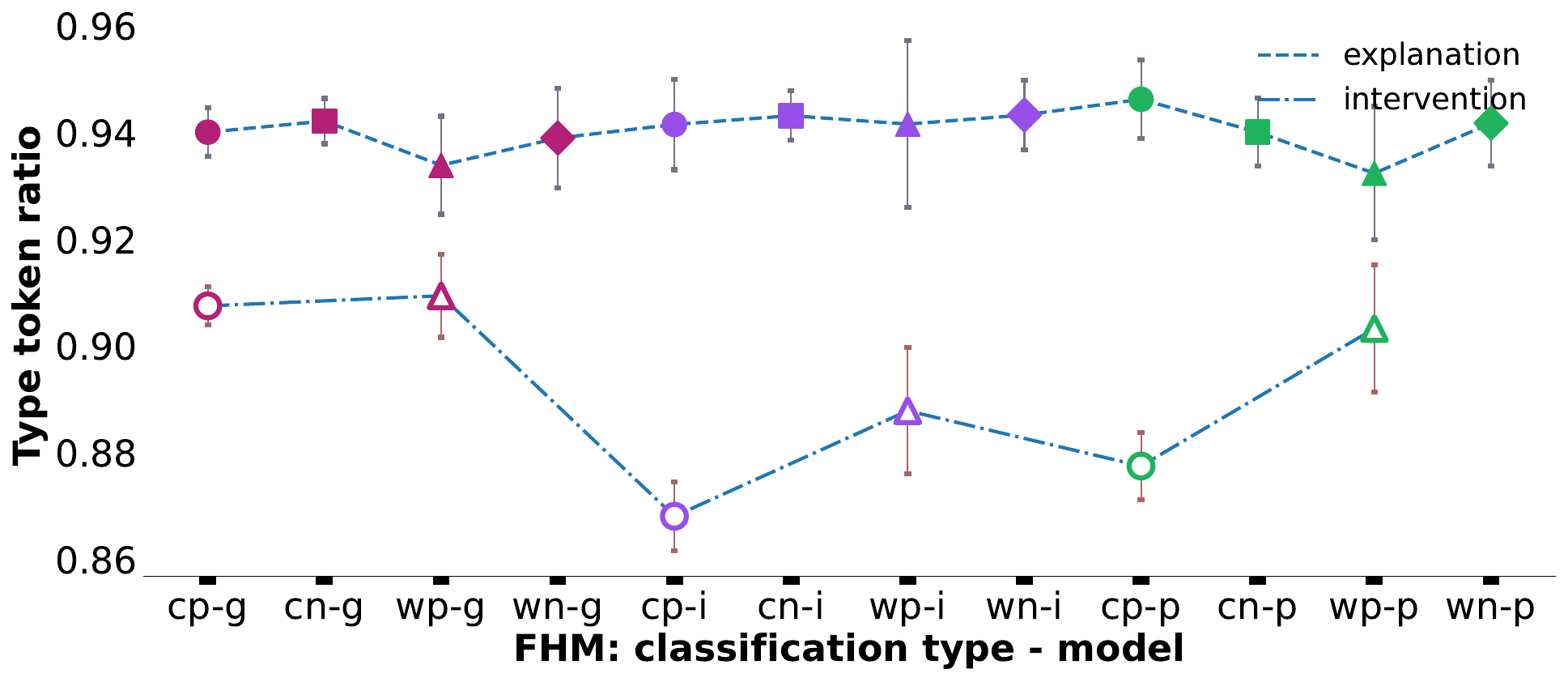}} &
\textbf{\includegraphics[width=0.33\linewidth]{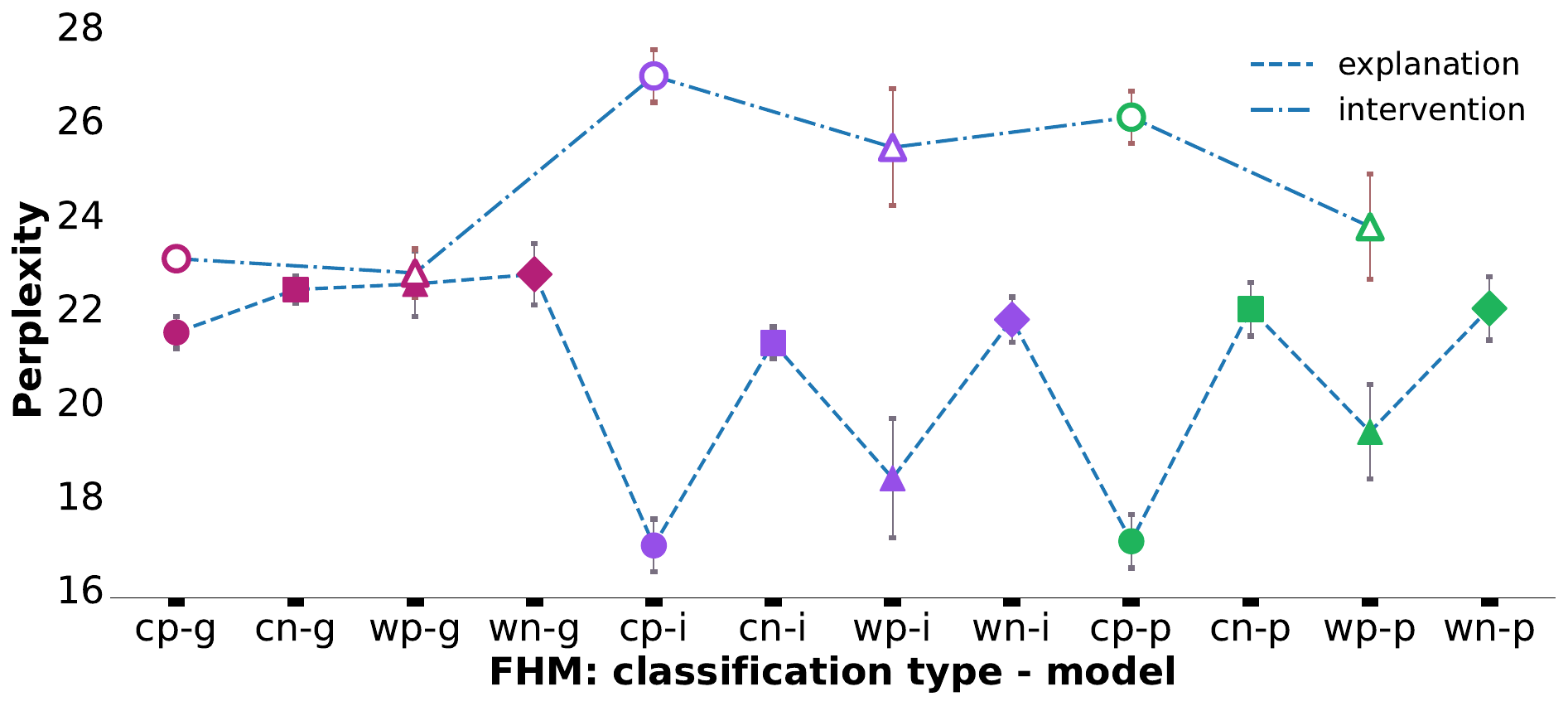}} \\ \\
\textbf{\includegraphics[width=0.33\linewidth]{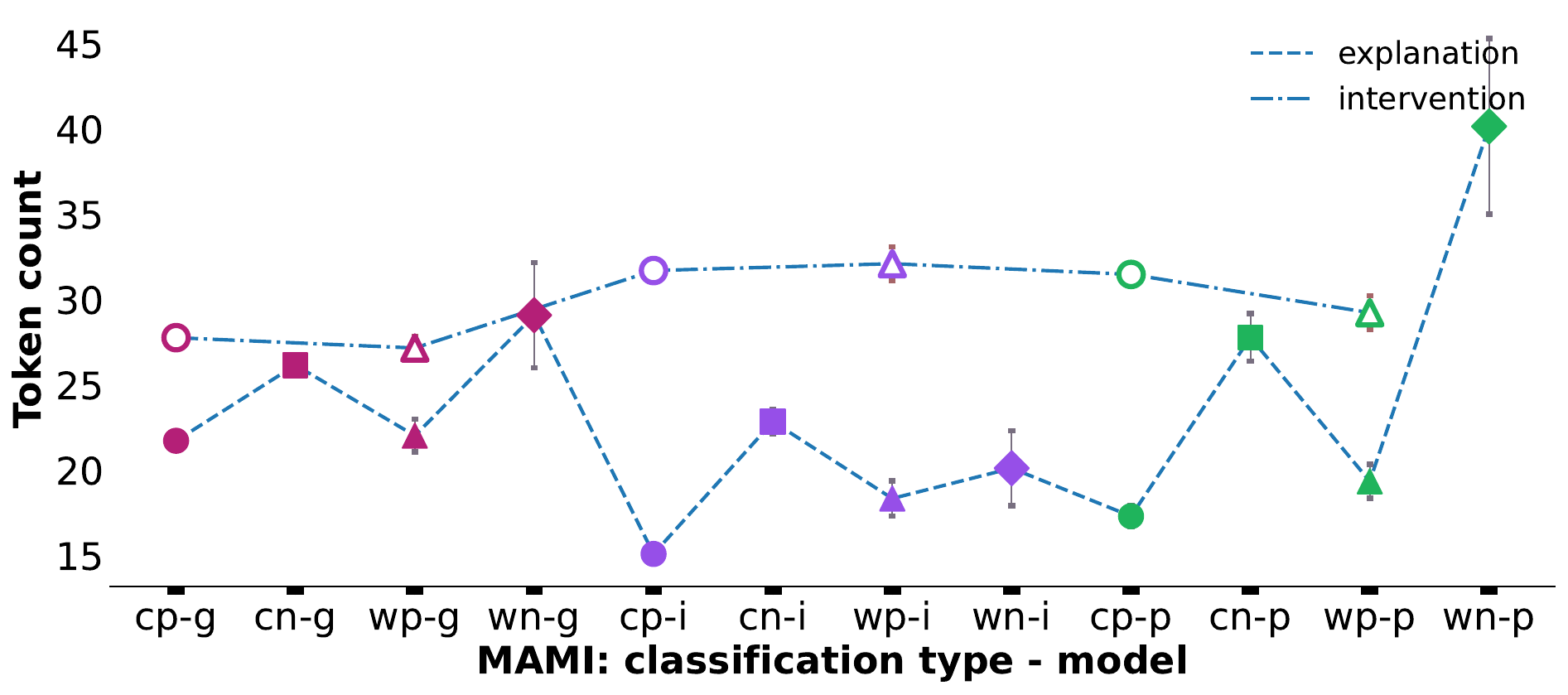}} &
\textbf{\includegraphics[width=0.33\linewidth]{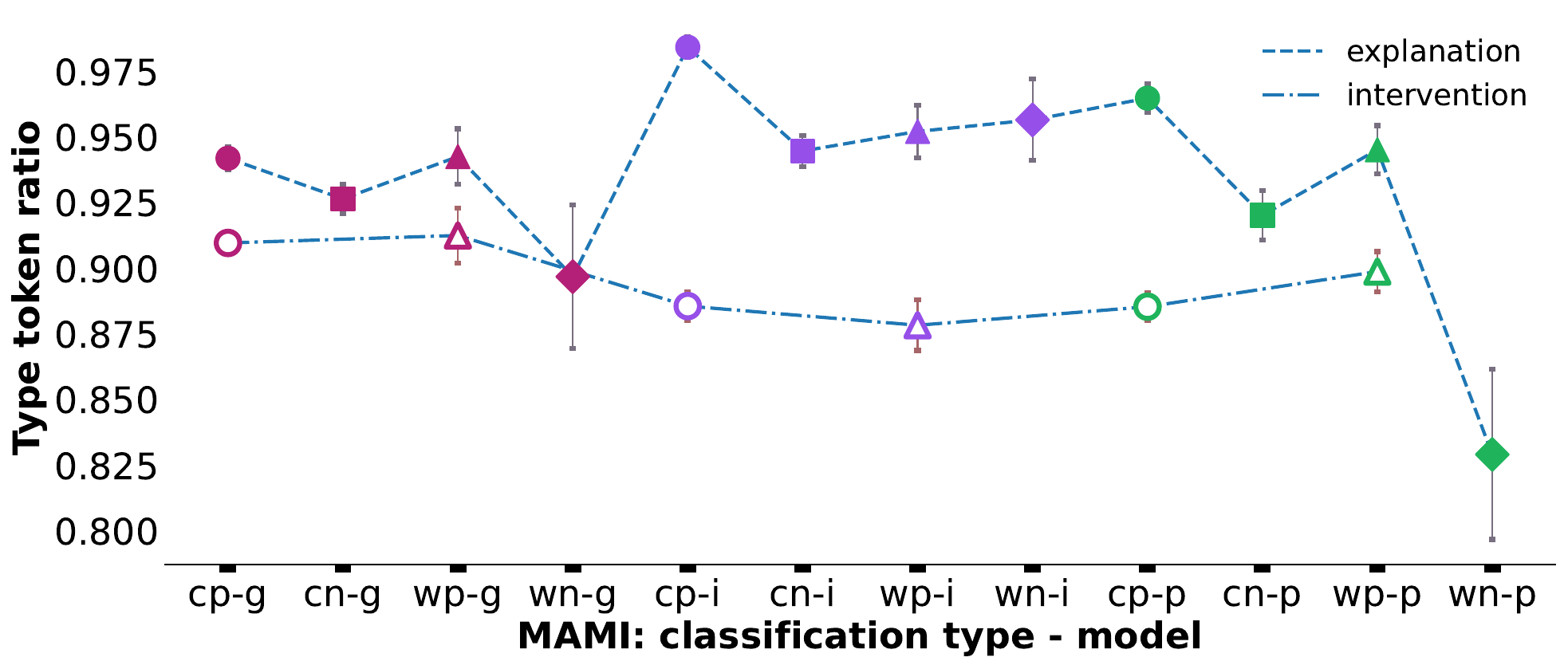}} &
\textbf{\includegraphics[width=0.33\linewidth]{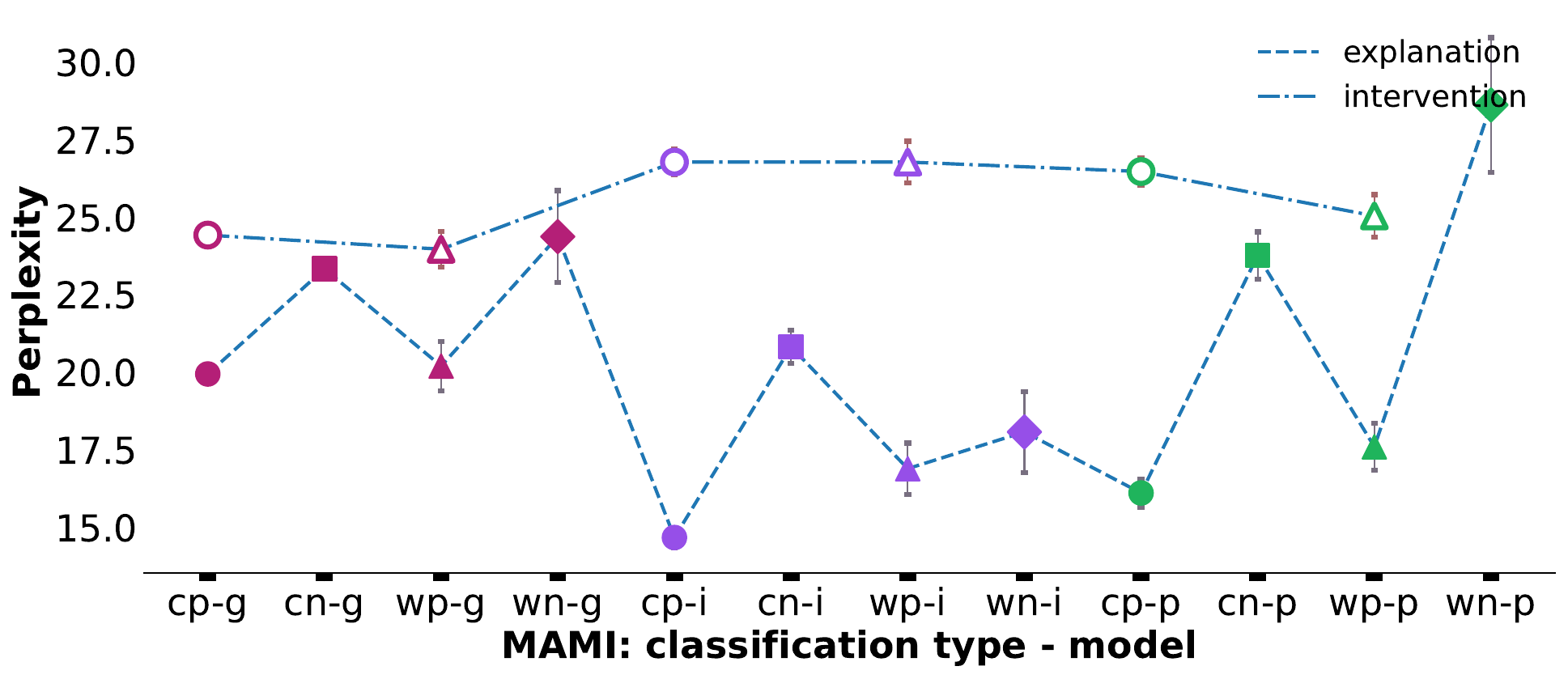}}
\end{tabular}
\captionof{figure}{\footnotesize Token count, type token ratio, and perplexity along with error bars at 95\% confidence interval. Here, \texttt{cp:} correct positive (circle), \texttt{cn:} correct negative (square), \texttt{wp:} wrong positive (triangle), \texttt{wn:} wrong negative (diamond), \texttt{g:} \sysGPT{} (maroon), \texttt{i:} \sysI{} (purple), and \texttt{p:} \sysPx{} (green).}
\label{fig:textual_analysis}
\end{table*}
One of the key novelties of our setup is the explanation and the intervention outputs, in addition to the classification label. In this section, we analyze the different textual properties of these outputs to demonstrate their characteristics. We measure each of these properties across the following subgroups -- correctly classified as hateful (\texttt{cp}), correctly classified as non-hateful (\texttt{cn}), incorrectly classified as hateful (\texttt{wp}) and incorrectly classified as non-hateful (\texttt{wn}). Note that while for explanation, all four subgroups are relevant, for intervention, only \texttt{cp} and \texttt{wp} are relevant.\\
\noindent\textit{Token count}: We count the number of tokens in the explanation and the intervention texts for each of the subgroups. Figure~\ref{fig:textual_analysis} shows that explanation texts are smaller than intervention texts across all subgroups, all models and both datasets. Next, for both explanation and intervention, the text length produced by \sysGPT{} is more consistent across all subgroups compared to the other models. This is a possible indication of the higher stability of this model over the others. Across the open models, the token count for the explanation of the non-hateful predictions (\texttt{cn} and \texttt{wn}) is larger than that of the hateful predictions (\texttt{cp} and \texttt{wp}), showing that the models present a larger explanation when an input is flagged as non-hateful.\\
\noindent\textit{Type-token ratio} (\textit{ttr}): The type-token ratio is a simple linguistic measure of vocabulary richness, calculated by dividing the number of \textit{unique} words (types) by the total number of words (tokens) in a text, showing how varied the vocabulary is. A higher \textit{ttr} means more diverse words, while a lower \textit{ttr} indicates more repetition. For all models, subgroups, and datasets the explanation text has a larger \textit{ttr} compared to the intervention text (Figure~\ref{fig:textual_analysis}). This indicates that while the models generate explanations with high lexical diversity, the intervention text is often repetitive. Taken together with the token count, the explanation text is short but more diverse while the intervention text is long but more repetitive.\\
\noindent\textit{Unigram perplexity}: The perplexity of the intervention text is higher than the explanation text across all subgroups, models and datasets (Figure~\ref{fig:textual_analysis}). This indicates that the explanation texts are linguistically more coherent than the intervention text. The reason could be two-fold -- (a) the fine-tuning data for intervention needs to be improved in future, (b) the models are better in generating explanations due to the recent developments in their reasoning capabilities right from the pretraining stage; intervention, on the other hand, is a relatively new concept and has possibly not been blended into the pretraining pipeline of most modern models. In addition, we also observe that the explanation text has a higher perplexity for the non-hateful classes (\texttt{cn} and \texttt{wn}), indicating that the models usually are less coherent in generating proper explanations while flagging a data point as non-hateful.\\
\if{0}\noindent\textit{Semantic relation}: Measuring the semantic relationship of the generated explanation and intervention in case where the model predicted the input to be hateful (i.e., \texttt{cp} and \texttt{wp}) is important to shed light on their `understanding' of hate speech. To this purpose, we measure the semantic similarity (discussed in prior Section~\ref{sec:results}) across all the models and datasets (Figure~\ref{fig:semantic_analysis}). \sysGPT{} has the highest semantic relationship over the other two models. Further, apart from the case of MAMI for \sysGPT{}, in all other cases, semantic similarity is higher for \texttt{wp} cases, signifying that the model strives to generate more appropriate explanation/intervention text to justify their wrong predictions.\\
\noindent\textit{Human validation}: As presented in Section~\ref{sec:results}, we perform manual evaluation here as well on \sysGPT{} and \sysI{}. We randomly sample 200 samples and pass these to the two annotators. Following are the obtained results: (i) \sysGPT{}: FHM -- (0.54, 0.556) for \texttt{cp}, (0.56, 0.564) for \texttt{wp}, MAMI -- (0.616, 0.604) for \texttt{cp}, (0.548, 0.58) for \texttt{wp} and (ii) \sysI{} -- (0.556, 0.572) for \texttt{cp}, (0.52, 0.576) for \texttt{wp}, MAMI -- (0.592, 0.62) for \texttt{cp}, (0.536, 0.588) for \texttt{wp}. All these values in brackets again correspond to two different annotators and are similar to what we discussed in Section~\ref{sec:results}. Overall these scores (and the observations therefore) match with the automated scores presented above.\\
\fi
\noindent\textit{Sentiment distribution}: To further understand the `tone' of the text generated by the models, we perform sentiment analysis over their generated explanation and intervention (refer Figure~\ref{fig:sentiment_analysis}). Each of the six plots contains six bars where the first four are for explanation and the remaining two are for intervention sentiments. Each bar contains the percentage distribution of three sentiments -- \textit{positive, neutral, negative} -- that are obtained by using \textsc{Vader} sentiment analysis tool\footnote{\url{https://github.com/cjhutto/vaderSentiment}}. The key observations based on the sentiment score are as follows -- \noindent(i) For explanation across both the datasets, cases where model performs positive predictions (i.e., \texttt{cp-ex} and \texttt{wp-ex}) have higher negative sentiment compared to negative prediction (i.e., \texttt{cn-ex} and \texttt{wn-ex}) that have higher positive sentiment. This observation indicates that models generally use negative (positive) sentiments to explain why an input should be flagged as hateful (non-hateful).
\noindent(ii) Intervention text (i.e., \texttt{cp-in} and \texttt{wp-in}) of \sysGPT{} has more positive sentiment in contrast to \sysI{} and \sysPx{}. This indicates that compared to \sysGPT{} the open models tend to better combat the hateful post with negatively charged intervention text and are thus found to align better with the ground truth (see Section~\ref{sec:results}).
\noindent(iii) The explanations for the MAMI data points that are predicted hateful have higher positive and neutral sentiments compared to the FHM data points. This indicates that models tend to elicit more positive sentiments to explain why an input is predicted as misogynistic. Further extended analysis is presented in the Appendix due to paucity of space.\\

\section{Ablation}

\subsection{Multi-agent synergy}
We conduct comprehensive 8-shot ablation studies using \sysGPT{} under multiple settings for FHM dataset: (i) a vanilla setup without any agent (\textsc{Vn}), (ii) single-agent variants where only $\mathcal{A}_\mathcal{C}$, $\mathcal{A}_\mathcal{E}$, or $\mathcal{A}_\mathcal{I}$ is employed, and (iii) alternative prompting paradigms including chain-of-thought (CoT) and zero-shot (ZS) prompting. Following are the obtained results: (i) Classification performance remains largely stable across ablations, with notable drops only for zero-shot (-4.36 mF1) and CoT (-3.55 mF1), highlighting the critical role of \sysSL{}-based few-shot retrieval. (ii) Explanation and intervention quality of different setups when compared against the complete three-agent framework ($\mathcal{A}_\mathcal{C}$+$\mathcal{A}_\mathcal{E}$+$\mathcal{A}_\mathcal{I}$) results in a substantial degradation. Specifically, the relative drops in semantic similarity (explanation, intervention) are: only $\mathcal{A}_\mathcal{C}$ (29\%, 28\%),  only $\mathcal{A}_\mathcal{E}$ (0\%, 28\%),  only $\mathcal{A}_\mathcal{I}$ (42\%, 3\%), \textsc{Vn} (34\%, 29\%), CoT (19\%, 25\%), and ZS (19\%, 26\%). These are large and consistent performance losses, demonstrating that neither individual agents nor conventional prompting approaches can effectively replicate the behavior of the full framework.

\subsection{Deployment cost}
We employ relatively small steering agents (with 3B parameters) that can be fully fine-tuned on commodity hardware (30 GB VRAM, batch size 1, bfloat16). At inference time, each agent requires nearly 10 GB VRAM, and, as demonstrated in in the main content, only the captioning agent needs to be deployed alongside the much larger black-box VLM \sysGPT{}. With 8-shot prompting, PEST with \sysGPT{} processes a sample in $\sim$ 6–7 seconds using an average of 1350 input tokens \& 50 output tokens, corresponding to a cost of roughly USD 0.0041 per sample (USD 4 for 1,000 samples).


\section{Conclusion}
\label{sec:conclusion}
In this paper, we present a first of its kind task of simultaneously classifying, explaining and intervening hateful memes and propose multiple datasets for training and evaluation purposes. Then to cater to the need of generalizability in the context of costly dataset curation, we propose a novel low-resource task specific agents based few-shot framework which we call \ps{} and provide various state-of-the-art results. Finally, we conclude with a detailed analysis of explanation and intervention texts generated by the models identifying their nuanced characteristics.


\section{Acknowledgements}
\label{sec:acknowledgements}
Naquee Rizwan dedicates this work in the memory of his loving late grandfather, \textbf{Dr. Mokhtar Hasan}, whose guidance and unwavering support have been a constant source of strength throughout his journey. AM acknowledges the funding from the ANRF-CRG.


\bibliographystyle{ACM-Reference-Format}
\bibliography{main}

@String{Computing = "Computing" }

@String{Springer = "Springer-Verlag" }

@ArtifactSoftware{R,
    title = {R: A Language and Environment for Statistical Computing},
    author = {{R Core Team}},
    organization = {R Foundation for Statistical Computing},
    address = {Vienna, Austria},
    year = {2019},
    url = {https://www.R-project.org/},
}

@article{huang2024towards,
  title={Towards low-resource harmful meme detection with LMM agents},
  author={Huang, Jianzhao and Lin, Hongzhan and Liu, Ziyan and Luo, Ziyang and Chen, Guang and Ma, Jing},
  journal={arXiv preprint arXiv:2411.05383},
  year={2024}
}

@article{Rizwan_Bhaskar_Das_Majhi_Saha_Mukherjee_2025, title={Exploring the Limits of Zero Shot Vision Language Models for Hate Meme Detection: The Vulnerabilities and their Interpretations}, volume={19}, url={https://ojs.aaai.org/index.php/ICWSM/article/view/35894}, DOI={10.1609/icwsm.v19i1.35894}, abstractNote={There is a rapid increase in the use of multimedia content in current social media platforms. One of the highly popular forms of such multimedia content are memes. While memes have been primarily invented to promote funny and buoyant discussions, malevolent users exploit memes to target individuals or vulnerable communities, making it imperative to identify and address such instances of hateful memes. Thus social media platforms are in dire need for active moderation of such harmful content. While manual moderation is extremely difficult due to the scale of such content, automatic moderation is challenged by the need of good quality annotated data to train hate meme detection algorithms. This makes a perfect pretext for exploring the power of modern day vision language models (VLMs) that have exhibited outstanding performance across various tasks. In this paper we study the effectiveness of VLMs in handling intricate tasks such as hate meme detection in a completely zero-shot setting so that there is no dependency on annotated data for the task. We perform thorough prompt engineering and query state-of-the-art VLMs using various prompt types to detect hateful/harmful memes. We further interpret the misclassification cases using a novel superpixel based occlusion method. Finally we show that these misclassifications can be neatly arranged into a typology of error classes the knowledge of which should enable the design of better safety guardrails in future. Code and other relevant sources are available online.
Warning: Contains potentially offensive content.}, number={1}, journal={Proceedings of the International AAAI Conference on Web and Social Media}, author={Rizwan, Naquee and Bhaskar, Paramananda and Das, Mithun and Majhi, Swadhin Satyaprakash and Saha, Punyajoy and Mukherjee, Animesh}, year={2025}, month={Jun.}, pages={1669-1689} }

@misc{swain2025toxictagsdecodingtoxicmemes,
      title={ToxicTAGS: Decoding Toxic Memes with Rich Tag Annotations}, 
      author={Subhankar Swain and Naquee Rizwan and Nayandeep Deb and Vishwajeet Singh Solanki and Vishwa Gangadhar S and Animesh Mukherjee},
      year={2025},
      eprint={2508.04166},
      archivePrefix={arXiv},
      primaryClass={cs.CV},
      url={https://arxiv.org/abs/2508.04166}, 
}

@article{Hee_Lee_2025, title={Demystifying Hateful Content: Leveraging Large Multimodal Models for Hateful Meme Detection with Explainable Decisions}, volume={19}, url={https://ojs.aaai.org/index.php/ICWSM/article/view/35845}, DOI={10.1609/icwsm.v19i1.35845}, abstractNote={Hateful meme detection presents a significant challenge as a multimodal task due to the complexity of interpreting implicit hate messages and contextual cues within memes. Previous approaches have fine-tuned pre-trained vision-language models (PT-VLMs), leveraging the knowledge they gained during pre-training and their attention mechanisms to understand meme content. However, the reliance of these models on implicit knowledge and complex attention mechanisms renders their decisions difficult to explain, which is crucial for building trust in meme classification. In this paper, we introduce IntMeme, a novel framework that leverages Large Multimodal Models (LMMs) for hateful meme classification with explainable decisions. IntMeme addresses the dual challenges of improving both accuracy and explainability in meme moderation. The framework uses LMMs to generate human-like, interpretive analyses of memes, providing deeper insights into multimodal content and context. Additionally, it uses independent encoding modules for both memes and their interpretations, which are then combined to enhance classification performance. Our approach addresses the opacity and misclassification issues associated with PT-VLMs, optimizing the use of LMMs for hateful meme detection. We demonstrate the effectiveness of IntMeme through comprehensive experiments across three datasets, showcasing its superiority over state-of-the-art models.}, number={1}, journal={Proceedings of the International AAAI Conference on Web and Social Media}, author={Hee, Ming Shan and Lee, Roy Ka-Wei}, year={2025}, month={Jun.}, pages={774-785} }

@misc{lin2024goatbench,
      title={GOAT-Bench: Safety Insights to Large Multimodal Models through Meme-Based Social Abuse}, 
      author={Hongzhan Lin and Ziyang Luo and Bo Wang and Ruichao Yang and Jing Ma},
      year={2024},
      eprint={2401.01523},
      archivePrefix={arXiv},
      primaryClass={cs.CL}
}

@misc{cao2023promptingmultimodalhatefulmeme,
      title={Prompting for Multimodal Hateful Meme Classification}, 
      author={Rui Cao and Roy Ka-Wei Lee and Wen-Haw Chong and Jing Jiang},
      year={2023},
      eprint={2302.04156},
      archivePrefix={arXiv},
      primaryClass={cs.CL},
      url={https://arxiv.org/abs/2302.04156}, 
}

@article{kiela2020hateful,
  title={The hateful memes challenge: Detecting hate speech in multimodal memes},
  author={Kiela, Douwe and Firooz, Hamed and Mohan, Aravind and Goswami, Vedanuj and Singh, Amanpreet and Ringshia, Pratik and Testuggine, Davide},
  journal={NeurIPS},
  volume={33},
  pages={2611--2624},
  year={2020}
}

@inproceedings{fersini2022semeval,
  title={SemEval-2022 Task 5: Multimedia automatic misogyny identification},
  author={Fersini, Elisabetta and Gasparini, Francesca and Rizzi, Giulia and Saibene, Aurora and Chulvi, Berta and Rosso, Paolo and Lees, Alyssa and Sorensen, Jeffrey},
  booktitle={Proceedings of the 16th International Workshop on Semantic Evaluation (SemEval-2022)},
  pages={533--549},
  year={2022}
}

@inproceedings{das2023banglaabusememe,
  title={BanglaAbuseMeme: A Dataset for Bengali Abusive Meme Classification},
  author={Das, Mithun and Mukherjee, Animesh},
  booktitle={Proceedings of the 2023 Conference on Empirical Methods in Natural Language Processing},
  pages={15498--15512},
  year={2023}
}

@inproceedings{10.1145/3485447.3512260,
author = {Hee, Ming Shan and Lee, Roy Ka-Wei and Chong, Wen-Haw},
title = {On Explaining Multimodal Hateful Meme Detection Models},
year = {2022},
isbn = {9781450390965},
publisher = {Association for Computing Machinery},
address = {New York, NY, USA},
url = {https://doi.org/10.1145/3485447.3512260},
doi = {10.1145/3485447.3512260},
booktitle = {Proceedings of the ACM Web Conference 2022},
pages = {3651–3655},
numpages = {5},
location = {<conf-loc>, <city>Virtual Event, Lyon</city>, <country>France</country>, </conf-loc>},
series = {WWW '22}
}

@inproceedings{lin2024towards,
  title={Towards explainable harmful meme detection through multimodal debate between large language models},
  author={Lin, Hongzhan and Luo, Ziyang and Gao, Wei and Ma, Jing and Wang, Bo and Yang, Ruichao},
  booktitle={Proceedings of the ACM Web Conference 2024},
  pages={2359--2370},
  year={2024}
}

@article{Hee2023DecodingTU,
  title={Decoding the Underlying Meaning of Multimodal Hateful Memes},
  author={Ming Shan Hee and Wen-Haw Chong and Roy Ka-Wei Lee},
  journal={ArXiv},
  year={2023},
  volume={abs/2305.17678},
  url={https://api.semanticscholar.org/CorpusID:258960556}
}

@article{kmainasi2025memeintel,
  title={MemeIntel: Explainable Detection of Propagandistic and Hateful Memes},
  author={Kmainasi, Mohamed Bayan and Hasnat, Abul and Hasan, Md Arid and Shahroor, Ali Ezzat and Alam, Firoj},
  journal={arXiv preprint arXiv:2502.16612},
  year={2025}
}

@misc{adak2025memesenseadaptiveincontextframework,
      title={MemeSense: An Adaptive In-Context Framework for Social Commonsense Driven Meme Moderation}, 
      author={Sayantan Adak and Somnath Banerjee and Rajarshi Mandal and Avik Halder and Sayan Layek and Rima Hazra and Animesh Mukherjee},
      year={2025},
      eprint={2502.11246},
      archivePrefix={arXiv},
      primaryClass={cs.IR},
      url={https://arxiv.org/abs/2502.11246}, 
}

@article{agarwal2024mememqa,
  title={MemeMQA: multimodal question answering for memes via rationale-based inferencing},
  author={Agarwal, Siddhant and Sharma, Shivam and Nakov, Preslav and Chakraborty, Tanmoy},
  journal={arXiv preprint arXiv:2405.11215},
  year={2024}
}

@article{hwang2023memecap,
  title={Memecap: A dataset for captioning and interpreting memes},
  author={Hwang, EunJeong and Shwartz, Vered},
  journal={arXiv preprint arXiv:2305.13703},
  year={2023}
}

@misc{cao2024modularizednetworksfewshothateful,
      title={Modularized Networks for Few-shot Hateful Meme Detection}, 
      author={Rui Cao and Roy Ka-Wei Lee and Jing Jiang},
      year={2024},
      eprint={2402.11845},
      archivePrefix={arXiv},
      primaryClass={cs.CL},
      url={https://arxiv.org/abs/2402.11845}, 
}

@inproceedings{hee-etal-2024-bridging,
    title = "Bridging Modalities: Enhancing Cross-Modality Hate Speech Detection with Few-Shot In-Context Learning",
    author = "Hee, Ming Shan  and
      Kumaresan, Aditi  and
      Lee, Roy Ka-Wei",
    editor = "Al-Onaizan, Yaser  and
      Bansal, Mohit  and
      Chen, Yun-Nung",
    booktitle = "Proceedings of the 2024 Conference on Empirical Methods in Natural Language Processing",
    month = nov,
    year = "2024",
    address = "Miami, Florida, USA",
    publisher = "Association for Computational Linguistics",
    url = "https://aclanthology.org/2024.emnlp-main.445/",
    doi = "10.18653/v1/2024.emnlp-main.445",
    pages = "7785--7799"
}

@article{PORTO2023100307,
title = {Usage of few-shot learning and meta-learning in agriculture: A literature review},
journal = {Smart Agricultural Technology},
volume = {5},
pages = {100307},
year = {2023},
issn = {2772-3755},
doi = {https://doi.org/10.1016/j.atech.2023.100307},
url = {https://www.sciencedirect.com/science/article/pii/S2772375523001363},
author = {João Vitor de Andrade Porto and Arlinda Cantero Dorsa and Vanessa Aparecida de Moraes Weber and Karla Rejane de Andrade Porto and Hemerson Pistori}
}

@inproceedings{10.1145/3675417.3675513,
author = {Li, Shiye and Yi, Li},
title = {A Few-Shot Entity Relation Extraction Method in the Legal Domain Based on Large Language Models},
year = {2024},
isbn = {9798400717147},
publisher = {Association for Computing Machinery},
address = {New York, NY, USA},
url = {https://doi.org/10.1145/3675417.3675513},
doi = {10.1145/3675417.3675513},
booktitle = {Proceedings of the 2024 Guangdong-Hong Kong-Macao Greater Bay Area International Conference on Digital Economy and Artificial Intelligence},
pages = {580–586},
numpages = {7},
location = {Hongkong, China},
series = {DEAI '24}
}

@misc{ayub2022fewshotcontinualactivelearning,
      title={Few-Shot Continual Active Learning by a Robot}, 
      author={Ali Ayub and Carter Fendley},
      year={2022},
      eprint={2210.04137},
      archivePrefix={arXiv},
      primaryClass={cs.LG},
      url={https://arxiv.org/abs/2210.04137}, 
}

@inproceedings{lin2004rouge,
  title={Rouge: A package for automatic evaluation of summaries},
  author={Lin, Chin-Yew},
  booktitle={Text summarization branches out},
  pages={74--81},
  year={2004}
}

@misc{zhang2020bertscoreevaluatingtextgeneration,
      title={BERTScore: Evaluating Text Generation with BERT}, 
      author={Tianyi Zhang and Varsha Kishore and Felix Wu and Kilian Q. Weinberger and Yoav Artzi},
      year={2020},
      eprint={1904.09675},
      archivePrefix={arXiv},
      primaryClass={cs.CL},
      url={https://arxiv.org/abs/1904.09675}, 
}

@article{jha2024memeguard,
  title={Memeguard: An llm and vlm-based framework for advancing content moderation via meme intervention},
  author={Jha, Prince and Jain, Raghav and Mandal, Konika and Chadha, Aman and Saha, Sriparna and Bhattacharyya, Pushpak},
  journal={arXiv preprint arXiv:2406.05344},
  year={2024}
}

@inproceedings{cao2023pro,
  title={Pro-cap: Leveraging a frozen vision-language model for hateful meme detection},
  author={Cao, Rui and Hee, Ming Shan and Kuek, Adriel and Chong, Wen-Haw and Lee, Roy Ka-Wei and Jiang, Jing},
  booktitle={Proceedings of the 31st ACM international conference on multimedia},
  pages={5244--5252},
  year={2023}
}

@article{hee2024bridging,
  title={Bridging modalities: Enhancing cross-modality hate speech detection with few-shot in-context learning},
  author={Hee, Ming Shan and Kumaresan, Aditi and Lee, Roy Ka-Wei},
  journal={arXiv preprint arXiv:2410.05600},
  year={2024}
}

@inproceedings{lu2025having,
  title={Is Having Rationales Enough? Rethinking Knowledge Enhancement for Multimodal Hateful Meme Detection},
  author={Lu, Junyu and Xu, Bo and Zhang, Xiaokun and Zhu, Haohao and Wang, Kaichun and Yang, Liang and Lin, Hongfei},
  booktitle={Proceedings of the 48th International ACM SIGIR Conference on Research and Development in Information Retrieval},
  pages={559--569},
  year={2025}
}

@article{Gallagher_2021,
   title={Generalized word shift graphs: a method for visualizing and explaining pairwise comparisons between texts},
   volume={10},
   ISSN={2193-1127},
   url={http://dx.doi.org/10.1140/epjds/s13688-021-00260-3},
   DOI={10.1140/epjds/s13688-021-00260-3},
   number={1},
   journal={EPJ Data Science},
   publisher={Springer Science and Business Media LLC},
   author={Gallagher, Ryan J. and Frank, Morgan R. and Mitchell, Lewis and Schwartz, Aaron J. and Reagan, Andrew J. and Danforth, Christopher M. and Dodds, Peter Sheridan},
   year={2021},
   month=jan }

@inproceedings{hee-etal-2024-recent,
    title = "Recent Advances in Online Hate Speech Moderation: Multimodality and the Role of Large Models",
    author = "Hee, Ming Shan  and
      Sharma, Shivam  and
      Cao, Rui  and
      Nandi, Palash  and
      Nakov, Preslav  and
      Chakraborty, Tanmoy  and
      Lee, Roy Ka-Wei",
    editor = "Al-Onaizan, Yaser  and
      Bansal, Mohit  and
      Chen, Yun-Nung",
    booktitle = "Findings of the Association for Computational Linguistics: EMNLP 2024",
    month = nov,
    year = "2024",
    address = "Miami, Florida, USA",
    publisher = "Association for Computational Linguistics",
    url = "https://aclanthology.org/2024.findings-emnlp.254/",
    doi = "10.18653/v1/2024.findings-emnlp.254",
    pages = "4407--4419"
}

@article{Mathew_Saha_Tharad_Rajgaria_Singhania_Maity_Goyal_Mukherjee_2019, title={Thou Shalt Not Hate: Countering Online Hate Speech}, volume={13}, url={https://ojs.aaai.org/index.php/ICWSM/article/view/3237}, DOI={10.1609/icwsm.v13i01.3237}, abstractNote={&lt;p&gt;Hate content in social media is ever increasing. While Facebook, Twitter, Google have attempted to take several steps to tackle the hateful content, they have mostly been unsuccessful. Counterspeech is seen as an effective way of tackling the online hate without any harm to the freedom of speech. Thus, an alternative strategy for these platforms could be to promote counterspeech as a defense against hate content. However, in order to have a successful promotion of such counterspeech, one has to have a deep understanding of its dynamics in the online world. Lack of carefully curated data largely inhibits such understanding. In this paper, we create and release the first ever dataset for counterspeech using comments from YouTube. The data contains 13,924 manually annotated comments where the labels indicate whether a comment is a counterspeech or not. This data allows us to perform a rigorous measurement study characterizing the linguistic structure of counterspeech for the first time. This analysis results in various interesting insights such as: the counterspeech comments receive much more likes as compared to the noncounterspeech comments, for certain communities majority of the non-counterspeech comments tend to be hate speech, the different types of counterspeech are not all equally effective and the language choice of users posting counterspeech is largely different from those posting non-counterspeech as revealed by a detailed psycholinguistic analysis. Finally, we build a set of machine learning models that are able to automatically detect counterspeech in YouTube videos with an F1-score of 0.71. We also build multilabel models that can detect different types of counterspeech in a comment with an F1-score of 0.60.&lt;/p&gt;}, number={01}, journal={Proceedings of the International AAAI Conference on Web and Social Media}, author={Mathew, Binny and Saha, Punyajoy and Tharad, Hardik and Rajgaria, Subham and Singhania, Prajwal and Maity, Suman Kalyan and Goyal, Pawan and Mukherjee, Animesh}, year={2019}, month={Jul.}, pages={369-380} }

@misc{pan2026readseeguidingunimodal,
      title={Read as You See: Guiding Unimodal LLMs for Low-Resource Explainable Harmful Meme Detection}, 
      author={Fengjun Pan and Xiaobao Wu and Tho Quan and Anh Tuan Luu},
      year={2026},
      eprint={2506.08477},
      archivePrefix={arXiv},
      primaryClass={cs.CL},
      url={https://arxiv.org/abs/2506.08477}, 
}

@inproceedings{ribeiro2016should,
  title={“Why Should I Trust You?”: Explaining the Predictions of Any Classifier},
  author={Ribeiro, Marco Tulio and Singh, Sameer and Guestrin, Carlos},
  booktitle={Proceedings KDD},
  year={2016},
  pages={1135--1144}
}

@inproceedings{selvaraju2017grad,
  title={Grad-CAM: Visual Explanations from Deep Networks via Gradient-based Localization},
  author={Selvaraju, Ramprasaath R. and Cogswell, Michael and Das, Abhishek and Vedantam, Ramakrishna and Parikh, Devi and Batra, Dhruv},
  booktitle={Proceedings of ICCV},
  year={2017},
  pages={618--626}
}

@inproceedings{Qian_ElSherief_Belding_Wang_2019,
  title={Learning to Generate Counter Speech for Online Hate Speech},
  author={Qian, Jing and ElSherief, Mai and Belding, Elizabeth and Wang, William Yang},
  booktitle={Proceedings EMNLP-IJCNLP},
  year={2019},
  pages={5577--5587}
}

@inproceedings{Kiela2020HatefulMemes,
  title={The Hateful Memes Challenge: Detecting Hate Speech in Multimodal Memes},
  author={Kiela, Douwe and Bhooshan, Hamed and Firooz, Hamed and Perez, Emilio and Testuggine, Davide and Patra, Barun and Vijayaraghavan, Prateek and Khot, Tushar and Zhu, Guanghui and Batra, Dhruv and others},
  booktitle={NeurIPS},
  year={2020}
}

@inproceedings{10.1145/3746027.3754553,
author = {Rizwan, Naquee and Deb, Nayandeep and Roy, Sarthak and Solanki, Vishwajeet Singh and Garimella, Kiran and Mukherjee, Animesh},
title = {Toxicity Begets Toxicity: Unraveling Conversational Chains in Political Podcasts},
year = {2025},
isbn = {9798400720352},
publisher = {Association for Computing Machinery},
address = {New York, NY, USA},
url = {https://doi.org/10.1145/3746027.3754553},
doi = {10.1145/3746027.3754553},
booktitle = {Proceedings of the 33rd ACM International Conference on Multimedia},
pages = {11776–11784},
numpages = {9},
location = {Dublin, Ireland},
series = {MM '25}
}

\appendix

\begin{table*}[!ht]
\footnotesize
\centering
\caption{\footnotesize Results on other embedding retrievers, i.e. on BLIP and CLIP. IVL: \sysI{}, PX: \sysPx{}, GPT: \sysGPT{}, BL: BLIP, and CL: CLIP. Best results across each model are marked \colorbox[HTML]{D9EAD3}{\textbf{green}} and second best results are marked \colorbox[HTML]{F1FCED}{\ul light green}.}
\setlength{\tabcolsep}{0.45mm}
\renewcommand{\arraystretch}{1.4}
\begin{tabular}{c|cc|cc|ccc|cccc|cc|ccc|cccc}
\rowcolor[HTML]{FFF2CC} 
\cellcolor[HTML]{D9D2E9} & \multicolumn{2}{c|}{\cellcolor[HTML]{D9D2E9}} & \multicolumn{2}{c|}{\cellcolor[HTML]{FFF2CC}\textbf{FHM-cls}} & \multicolumn{3}{c|}{\cellcolor[HTML]{FFF2CC}\textbf{FHM-exp}} & \multicolumn{4}{c|}{\cellcolor[HTML]{FFF2CC}\textbf{FHM-int}} & \multicolumn{2}{c|}{\cellcolor[HTML]{FFF2CC}\textbf{MAMI-cls}} & \multicolumn{3}{c|}{\cellcolor[HTML]{FFF2CC}\textbf{MAMI-exp}} & \multicolumn{4}{c}{\cellcolor[HTML]{FFF2CC}\textbf{MAMI-int}} \\ \cline{4-21} 
\rowcolor[HTML]{FCE5CD} 
\multirow{-2}{*}{\cellcolor[HTML]{D9D2E9}\textbf{model}} & \multicolumn{2}{c|}{\multirow{-2}{*}{\cellcolor[HTML]{D9D2E9}\textbf{shots}}} & \textbf{acc} & \textbf{mf1} & \textbf{rgL} & \textbf{ss} & \textbf{bsf1} & \textbf{rgL} & \textbf{ss} & \textbf{bsf1} & \textbf{size} & \textbf{acc} & \textbf{mf1} & \textbf{rgL} & \textbf{ss} & \textbf{bsf1} & \textbf{rgL} & \textbf{ss} & \textbf{bsf1} & \textbf{size} \\ \hline

& \multicolumn{1}{c|}{} & \cellcolor[HTML]{D9D2E9}\textbf{2} & 78.23 & 78.23 & 0.221 & 0.652 & 0.886 & 0.144 & 0.466 & 0.872 & 389 & 86.35 & 86.22 & 0.229 & 0.583 & 0.882 & 0.127 & 0.435 & 0.864 & 454 \\
& \multicolumn{1}{c|}{} & \cellcolor[HTML]{D9D2E9}\textbf{4} & 78.54 & 78.53 & 0.233 & 0.671 & 0.889 & 0.164 & 0.509 & 0.876 & 393 & 86.03 & 85.90 & 0.230 & \cellcolor[HTML]{F1FCED}{\ul 0.615} & \cellcolor[HTML]{F1FCED}{\ul 0.884} & 0.126 & 0.480 & 0.866 & 452 \\
& \multicolumn{1}{c|}{\multirow{-3}{*}{\textbf{BL}}} & \cellcolor[HTML]{D9D2E9}\textbf{8} & \cellcolor[HTML]{D9EAD3}\textbf{79.96} & \cellcolor[HTML]{D9EAD3}\textbf{79.95} & \cellcolor[HTML]{F1FCED}{\ul 0.245} & \cellcolor[HTML]{D9EAD3}\textbf{0.684} & \cellcolor[HTML]{D9EAD3}\textbf{0.892} & \cellcolor[HTML]{D9EAD3}\textbf{0.207} & \cellcolor[HTML]{D9EAD3}\textbf{0.576} & \cellcolor[HTML]{D9EAD3}\textbf{0.886} & 402 & 88.47 & 88.42 & \cellcolor[HTML]{D9EAD3}\textbf{0.241} & \cellcolor[HTML]{D9EAD3}\textbf{0.622} & \cellcolor[HTML]{D9EAD3}\textbf{0.885} & 0.141 & 0.487 & 0.869 & 447 \\ \cline{2-21}
& \multicolumn{1}{c|}{} & \cellcolor[HTML]{D9D2E9}\textbf{2} & \cellcolor[HTML]{F1FCED}{\ul 79.45} & \cellcolor[HTML]{F1FCED}{\ul 79.44} & 0.220 & 0.654 & 0.886 & 0.144 & 0.467 & 0.871 & 403 & 88.47 & 88.43 & 0.219 & 0.582 & 0.880 & 0.139 & 0.508 & 0.870 & 443 \\
& \multicolumn{1}{c|}{} & \cellcolor[HTML]{D9D2E9}\textbf{4} & 78.94 & 78.93 & 0.237 & \cellcolor[HTML]{F1FCED}{\ul 0.678} & \cellcolor[HTML]{F1FCED}{\ul 0.890} & 0.167 & 0.495 & 0.875 & 400 & \cellcolor[HTML]{F1FCED}{\ul 88.99} & \cellcolor[HTML]{F1FCED}{\ul 88.97} & 0.230 & 0.602 & 0.882 & \cellcolor[HTML]{F1FCED}{\ul 0.179} & \cellcolor[HTML]{F1FCED}{\ul 0.585} & \cellcolor[HTML]{F1FCED}{\ul 0.879} & 444 \\
\multirow{-6}{*}{\textbf{GPT}} & \multicolumn{1}{c|}{\multirow{-3}{*}{\textbf{CL}}} & \cellcolor[HTML]{D9D2E9}\textbf{8} & 78.64 & 78.63 & \cellcolor[HTML]{D9EAD3}\textbf{0.246} & 0.683 & \cellcolor[HTML]{D9EAD3}\textbf{0.892} & \cellcolor[HTML]{F1FCED}{\ul 0.201} & \cellcolor[HTML]{F1FCED}{\ul 0.565} & \cellcolor[HTML]{F1FCED}{\ul 0.884} & 395 & \cellcolor[HTML]{D9EAD3}\textbf{89.63} & \cellcolor[HTML]{D9EAD3}\textbf{89.60} & \cellcolor[HTML]{F1FCED}{\ul 0.237} & 0.611 & 0.883 & \cellcolor[HTML]{D9EAD3}\textbf{0.244} & \cellcolor[HTML]{D9EAD3}\textbf{0.701} & \cellcolor[HTML]{D9EAD3}\textbf{0.894} & 450 \\ \hline

& \multicolumn{1}{c|}{} & \cellcolor[HTML]{D9D2E9}\textbf{2} & \cellcolor[HTML]{F1FCED}{\ul 73.45} & \cellcolor[HTML]{F1FCED}{\ul 73.42} & 0.206 & 0.601 & 0.878 & 0.229 & 0.620 & 0.892 & 336 & 74.60 & 74.47 & \cellcolor[HTML]{D9EAD3}\textbf{0.200} & \cellcolor[HTML]{F1FCED}{\ul 0.545} & \cellcolor[HTML]{F1FCED}{\ul 0.878} & 0.332 & 0.786 & 0.914 & 387 \\
& \multicolumn{1}{c|}{} & \cellcolor[HTML]{D9D2E9}\textbf{4} & 73.25 & 73.02 & 0.210 & 0.620 & 0.880 & 0.272 & 0.729 & \cellcolor[HTML]{F1FCED}{\ul 0.904} & 311 & 75.45 & 75.38 & \cellcolor[HTML]{F1FCED}{\ul 0.180} & 0.534 & 0.875 & 0.375 & 0.839 & \cellcolor[HTML]{D9EAD3}\textbf{0.920} & 381 \\
& \multicolumn{1}{c|}{\multirow{-3}{*}{\textbf{BL}}} & \cellcolor[HTML]{D9D2E9}\textbf{8} & 71.21 & 70.07 & \cellcolor[HTML]{F1FCED}{\ul 0.225} & \cellcolor[HTML]{F1FCED}{\ul 0.630} & 0.882 & \cellcolor[HTML]{F1FCED}{\ul 0.285} & \cellcolor[HTML]{D9EAD3}\textbf{0.788} & \cellcolor[HTML]{D9EAD3}\textbf{0.907} & 248 & \cellcolor[HTML]{F1FCED}{\ul 77.25} & \cellcolor[HTML]{F1FCED}{\ul 77.18} & 0.165 & 0.524 & 0.872 & \cellcolor[HTML]{D9EAD3}\textbf{0.384} & \cellcolor[HTML]{F1FCED}{\ul 0.844} & \cellcolor[HTML]{F1FCED}{\ul 0.919} & 390 \\ \cline{2-21}
& \multicolumn{1}{c|}{} & \cellcolor[HTML]{D9D2E9}\textbf{2} & 71.21 & 71.20 & 0.212 & 0.606 & 0.883 & 0.216 & 0.583 & 0.889 & 339 & 73.44 & 73.17 & \cellcolor[HTML]{D9EAD3}\textbf{0.200} & \cellcolor[HTML]{D9EAD3}\textbf{0.550} & \cellcolor[HTML]{D9EAD3}\textbf{0.879} & 0.326 & 0.786 & 0.914 & 394 \\
& \multicolumn{1}{c|}{} & \cellcolor[HTML]{D9D2E9}\textbf{4} & \cellcolor[HTML]{D9EAD3}\textbf{73.55} & \cellcolor[HTML]{D9EAD3}\textbf{73.44} & 0.222 & \cellcolor[HTML]{F1FCED}{\ul 0.630} & \cellcolor[HTML]{F1FCED}{\ul 0.885} & 0.271 & 0.723 & 0.903 & 323 & 74.50 & 74.39 & \cellcolor[HTML]{F1FCED}{\ul 0.180} & 0.538 & 0.875 & 0.377 & 0.836 & \cellcolor[HTML]{D9EAD3}\textbf{0.920} & 383 \\
\multirow{-6}{*}{\textbf{IVL}} & \multicolumn{1}{c|}{\multirow{-3}{*}{\textbf{CL}}} & \cellcolor[HTML]{D9D2E9}\textbf{8} & 70.60 & 69.51 & \cellcolor[HTML]{D9EAD3}\textbf{0.233} & \cellcolor[HTML]{D9EAD3}\textbf{0.641} & \cellcolor[HTML]{D9EAD3}\textbf{0.889} & \cellcolor[HTML]{D9EAD3}\textbf{0.290} & \cellcolor[HTML]{F1FCED}{\ul 0.778} & \cellcolor[HTML]{D9EAD3}\textbf{0.907} & 251 & \cellcolor[HTML]{D9EAD3}\textbf{77.67} & \cellcolor[HTML]{D9EAD3}\textbf{77.64} & 0.161 & 0.522 & 0.871 & \cellcolor[HTML]{F1FCED}{\ul 0.381} & \cellcolor[HTML]{D9EAD3}\textbf{0.846} & \cellcolor[HTML]{F1FCED}{\ul 0.919} & 386 \\ \hline

& \multicolumn{1}{c|}{} & \cellcolor[HTML]{D9D2E9}\textbf{2} & 69.79 & 69.78 & 0.191 & 0.559 & 0.879 & 0.223 & 0.634 & 0.889 & 339 & 68.78 & 67.23 & 0.191 & 0.488 & 0.877 & 0.327 & 0.798 & 0.912 & 428 \\
& \multicolumn{1}{c|}{} & \cellcolor[HTML]{D9D2E9}\textbf{4} & \cellcolor[HTML]{F1FCED}{\ul 71.52} & \cellcolor[HTML]{F1FCED}{\ul 71.51} & 0.206 & 0.598 & 0.882 & 0.255 & 0.718 & 0.898 & 343 & 73.94 & 73.44 & \cellcolor[HTML]{D9EAD3}\textbf{0.196} & 0.510 & 0.877 & 0.355 & 0.832 & \cellcolor[HTML]{F1FCED}{\ul 0.918} & 413 \\
& \multicolumn{1}{c|}{\multirow{-3}{*}{\textbf{BL}}} & \cellcolor[HTML]{D9D2E9}\textbf{8} & \cellcolor[HTML]{D9EAD3}\textbf{73.04} & \cellcolor[HTML]{D9EAD3}\textbf{72.95} & \cellcolor[HTML]{F1FCED}{\ul 0.218} & \cellcolor[HTML]{F1FCED}{\ul 0.625} & \cellcolor[HTML]{F1FCED}{\ul 0.885} & \cellcolor[HTML]{D9EAD3}\textbf{0.275} & \cellcolor[HTML]{D9EAD3}\textbf{0.771} & \cellcolor[HTML]{D9EAD3}\textbf{0.905} & 329 & \cellcolor[HTML]{D9EAD3}\textbf{76.38} & \cellcolor[HTML]{D9EAD3}\textbf{76.13} & 0.170 & \cellcolor[HTML]{F1FCED}{\ul 0.522} & 0.873 & \cellcolor[HTML]{D9EAD3}\textbf{0.389} & \cellcolor[HTML]{D9EAD3}\textbf{0.847} & \cellcolor[HTML]{D9EAD3}\textbf{0.920} & 409 \\ \cline{2-21}
& \multicolumn{1}{c|}{} & \cellcolor[HTML]{D9D2E9}\textbf{2} & 69.18 & 69.18 & 0.191 & 0.551 & 0.878 & 0.207 & 0.608 & 0.886 & 341 & 64.55 & 61.55 & \cellcolor[HTML]{F1FCED}{\ul 0.193} & 0.503 & \cellcolor[HTML]{D9EAD3}\textbf{0.880} & 0.332 & 0.805 & 0.912 & 437 \\
& \multicolumn{1}{c|}{} & \cellcolor[HTML]{D9D2E9}\textbf{4} & 70.70 & 70.68 & 0.213 & 0.603 & 0.884 & 0.250 & 0.712 & 0.897 & 333 & 69.65 & 68.49 & 0.192 & 0.510 & \cellcolor[HTML]{F1FCED}{\ul 0.878} & 0.359 & \cellcolor[HTML]{F1FCED}{\ul 0.835} & \cellcolor[HTML]{F1FCED}{\ul 0.918} & 417 \\
\multirow{-6}{*}{\textbf{PX}} & \multicolumn{1}{c|}{\multirow{-3}{*}{\textbf{CL}}} & \cellcolor[HTML]{D9D2E9}\textbf{8} & 71.31 & 71.31 & \cellcolor[HTML]{D9EAD3}\textbf{0.222} & \cellcolor[HTML]{D9EAD3}\textbf{0.631} & \cellcolor[HTML]{D9EAD3}\textbf{0.886} & \cellcolor[HTML]{F1FCED}{\ul 0.268} & \cellcolor[HTML]{F1FCED}{\ul 0.759} & \cellcolor[HTML]{F1FCED}{\ul 0.903} & 321 & \cellcolor[HTML]{F1FCED}{\ul 74.29} & \cellcolor[HTML]{F1FCED}{\ul 73.81} & 0.169 & \cellcolor[HTML]{D9EAD3}\textbf{0.523} & 0.873 & \cellcolor[HTML]{F1FCED}{\ul 0.368} & \cellcolor[HTML]{D9EAD3}\textbf{0.847} & \cellcolor[HTML]{F1FCED}{\ul 0.918} & 415 \\ \hline

\end{tabular}
\label{tab:few_shot_results}
\end{table*}


\section{Limitations}
\label{sec:limitations}
Our dataset is primarily in English and therefore the effectiveness of the helper agents needs to be checked for other languages in future. While compute and data are both limited for full fine-tuning, in future, upon availability of these resources an interesting study would be to compare large fine-tuned models with our few-shot setup across all three axes (classification, explanation, and intervention).


\section{Ethics statement}
\label{sec:ethics_statement}
Throughout this work, we did not intend to disclose the private details of users associated with the data. We have extended previous datasets for explanation and intervention without the addition of new samples, and therefore have taken proper care of privacy as per the corresponding copyright claims of the actual dataset providers. We also run our experiments with multiple models including both-- open-source \& proprietary -- to ensure robust conclusions. Lastly, we shall release our dataset upon acceptance of the paper while taking proper care so that there is no privacy leakage.


\section{Experimental setup}
\label{app:experimental_setup}
\subsection{Finetuning agents}
We fine-tune all agents using \sysP{} as it suitably fits on our 48GB-L40 server. The model used is openly available at HuggingFace (\texttt{google/paligemma-3b-pt-448}) that has all the relevant APIs for supervised fine-tuning. We employed \texttt{learning\_rate} of 2e-5 with Adam optimizer and a \texttt{weight\_decay} of 1e-6 for 2 epochs. At both stages (fine-tune and inference) and for all task specific agents, we use the standard bfloat-16 representation for weights and bias representation.

\subsection{Inference}
For all inference experiments, we run the models on a very low temperature value of 0.001 to ensure reproducibility. The \texttt{max\_new\_tokens} parameter is tuned to 100 tokens. For \sysGPT{}, we use the Microsoft Azure services and deploy the open-source models \sysI{}, \sysPx{}, and \sysP{} on our server with bfloat-16 representation. For \sysI{} and \sysPx{} we use their HuggingFace checkpoints.


\section{Other embedding retrievers}
\label{app:embed_retrievers}

In this section, we present the results of \ps{} for setups where in-context exemplar selection is based on CLIP and BLIP embedding similarity. Table~\ref{tab:few_shot_results} presents the comprehensive results across both FHM and MAMI datasets using BLIP and CLIP vision encoders for few-shot exemplar selection with 2, 4, and 8 shots. We make the following observations:

\noindent\textit{Results aligned to \sysSL{}}: We observe similar trends across the three types of tasks as presented with \sysSL{} embeddings in the main content. The best classification, explanation, and intervention scores are nearly matching for all three embedding retrievers with slight differences. Here as well \sysGPT{} performs the best on classification and explanation tasks, and intervention performs the best with the open-source model (i.e. \sysI{} and \sysPx{}). This points to an important conclusion that any effective visual embedding retriever module is able to exploit the best capability of the deployed multi-tasking inference model.

\noindent\textit{Shot scaling behavior}: Across all encoders, increasing the number of few-shot examples generally improves performance, particularly for \sysGPT{} and \sysPx{} which shows consistently increasing or nearly same scores for 2, 4 and 8 shots. However, \sysI{} exhibits slight non-monotonic behavior, suggesting potential sensitivity of the model towards the shot examples.

\noindent While all the retrievers worked well, \sysSL{} proved to be the most consistent, exhibiting the best results. Therefore, we presented these results in the main section.


\section{Prompts and definitions}
\label{app:prompts}
\subsection{\textbf{Deployed prompts}}

The exact prompt for generating the responses using the \ps{} framework is as follows. We pick the corresponding definitions of FHM/MAMI datasets from previous work~\cite{Rizwan_Bhaskar_Das_Majhi_Saha_Mukherjee_2025} and these are noted in the upcoming sub-section.

\begin{tcolorbox}[colback=blue!5!white,colframe=blue!75!black,title=System prompt]
\footnotesize
You are an AI assistant tasked with classifying memes into hateful or non-hateful along with an explanation while also generating an intervention only for hateful memes based on the input image, image caption and extracted text obtained from the image.\\
Consider the following definitions.\\
$\bullet$~hateful: \{definition of hateful\}\\
$\bullet$~non-hateful: \{definition of non-hateful\}\\

Classify the meme as hateful or non-hateful and then explain the reason based on the above definitions considering the image, provided caption of the image, as well as the extracted text from the image delimited by three backticks. If the meme is hateful then further generate an intervention for it.\\

Example output for hateful meme: \{hateful -- explanation in 30 words that why the meme is hateful. Intervention - intervention in 30 words for the hateful meme.\}\\
Example output for non-hateful meme: \{non-hateful - explanation in 30 words that why the meme is non-hateful.\}
\end{tcolorbox}
\begin{tcolorbox}[colback=blue!5!white,colframe=blue!75!black,title=User/Assistant prompt (2-shots drawn from $\mathcal{T}_\mathcal{R}$)]
\footnotesize
\textbf{User}:<image-1> \texttt{caption\_1} of the first shot image. OCR text of the first shot image.\\
\textbf{Assistant}: hateful - \texttt{explanation\_1} - \texttt{intervention\_1}\\ \\
\textbf{User}:<image-2> \texttt{caption\_2} of the second shot image. OCR text of the second shot image.\\
\textbf{Assistant}: non-hateful - \texttt{explanation\_2} 
\end{tcolorbox}
\noindent In the above prompt $\mathcal{A}_\mathcal{C}$(image-1) = \texttt{caption\_1}, $\mathcal{A}_\mathcal{E}$(image-1) = \texttt{explanation\_1}, and $\mathcal{A}_\mathcal{I}$(image-1) = \texttt{intervention\_1}. Similarly, $\mathcal{A}_\mathcal{C}$(image-2) = \texttt{caption\_2} and $\mathcal{A}_\mathcal{E}$(image-2) = \texttt{explanation\_2}.

\begin{tcolorbox}[colback=blue!5!white,colframe=blue!75!black,title=Test point drawn from $\mathcal{T}_\mathcal{S}$]
\footnotesize
\textbf{User}:<image-test> \texttt{caption\_test} of the test image. OCR text of the test image.\\
\textbf{Assistant}:
\end{tcolorbox}
\noindent In the above prompt $\mathcal{A}_\mathcal{C}$(image-test) = \texttt{caption\_test}.

\noindent The following definitions are drawn from prior works~\cite{kiela2020hateful, fersini2022semeval}.\\

\subsection{Definitions}

\noindent\textbf{FHM dataset}
\begin{itemize}
    \item \textbf{hateful}: A direct or indirect attack on people based on
characteristics, including ethnicity, race, nationality, immigration status, religion, caste, sex, gender identity, sexual orientation, and disability or disease. Attack is defined as violent or dehumanizing (comparing people to non-human things, e.g., animals) speech, statements of inferiority, and calls for exclusion or segregation. Mocking hate crime is also considered hateful.
    \item \textbf{non-hateful}: A meme which is not hateful and follows social norms.
\end{itemize}

\noindent\textbf{MAMI dataset}
\begin{itemize}
    \item \textbf{misogynistic}: A meme is misogynous if it conceptually describes an offensive, sexist or hateful scene (weak or strong, implicitly or explicitly) having as target a woman or a group of women. Misogyny can be expressed in the form of shaming, stereotype, objectification and/or violence.
    \item \textbf{non-misogynistic}: A meme that does not express any form of hate against women.
\end{itemize}


\begin{table*}[!t]
\centering
\footnotesize
\setlength{\tabcolsep}{0.01mm}
\begin{tabular}{ccc}
\textbf{\includegraphics[width=0.32\linewidth]{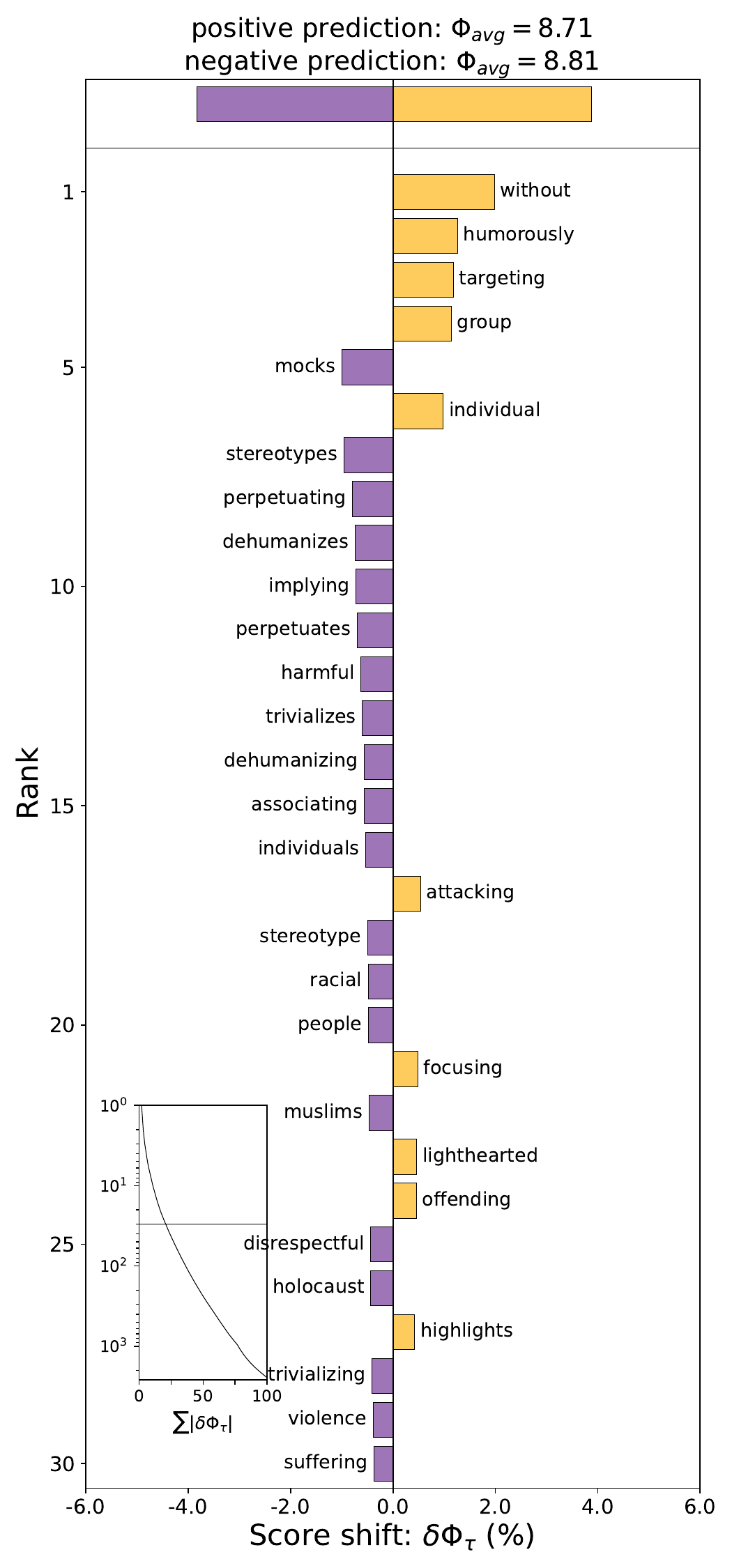}} &
\textbf{\includegraphics[width=0.32\linewidth]{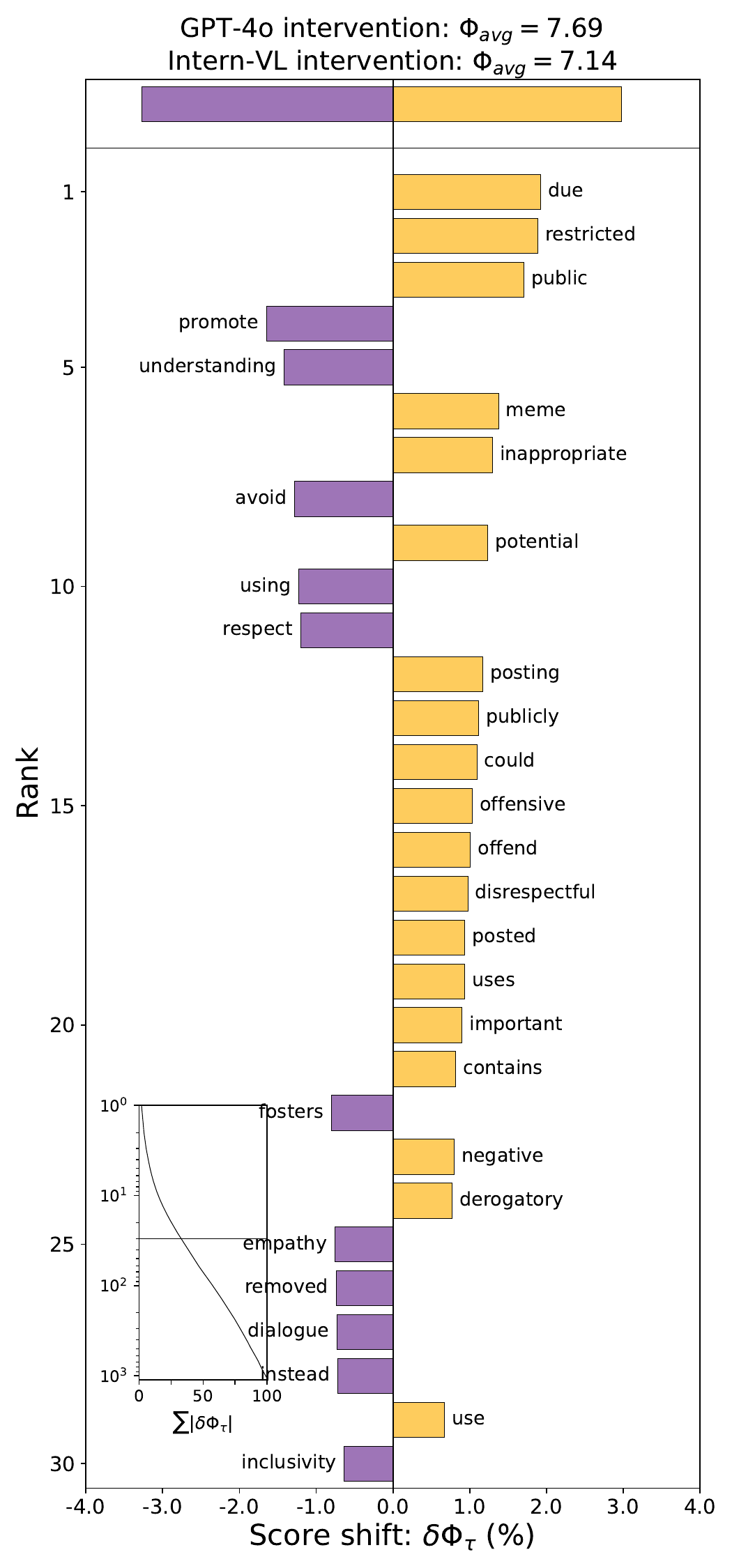}}\\
\texttt{(i)} &
\texttt{(ii)}
\end{tabular}
\captionof{figure}{\footnotesize Word shift graphs computed using Shannon entropy shifts. Each of the two word shift figures presents 30 ranked keywords based on entropy difference among the classes, with a cumulative entropy shift distribution graph present at the bottom left corner.}
\label{fig:word_shift_graph}
\end{table*}


\begin{table}[]
\footnotesize
\centering
\caption{Obtained $p$-values of exact two-sided binomial \textsc{McNemar} test on two different setups. The upper block presents the comparison of results on \sysGPT{} with 8 shots vs (a) U-CoT+, and (b) corresponding 8-shot results of \sysI{} and \sysPx{}. The lower block presents a comparison of employed VLMs in the 8-shot setup with their zero-shot setup. Number in \texttt{(parenthesis)} denotes the number of shots employed in the VLM.}
\setlength{\tabcolsep}{0.9mm}
\renewcommand{\arraystretch}{1.75}
\begin{tabular}{c|c|cc}
\textbf{setup-1} & \textbf{setup-2} & \textbf{FHM} & \textbf{MAMI} \\ \hline \hline
\multirow{3}{*}{\sysGPT{} \texttt{(8)}} & U-CoT+ & $2.7 \times 10^{-5}$ & $4.4 \times 10^{-15}$ \\
 & \sysI{} \texttt{(8)} & $6.1 \times 10^{-8}$ & $1.1 \times 10^{-20}$ \\
 & \sysPx{} \texttt{(8)} & $2.9 \times 10^{-7}$ & $3.8 \times 10^{-21}$ \\ \hline
\sysGPT{} \texttt{(8)} & \sysGPT{} \texttt{(0)} & $4.5 \times 10^{-4}$ & 0.01 \\
\sysI{} \texttt{(8)} & \sysI{} \texttt{(0)} & 0.051 & $7.7 \times 10^{-5}$ \\
\sysPx{} \texttt{(8)} & \sysPx{} \texttt{(0)} & 0.003 & 0.404
\end{tabular}
\label{tab:significance_test}
\end{table}

\section{Significance test}
\label{app:significance}

We have empirically shown that \ps{} outperforms various prior classification frameworks by significant margins in the main content. Here, we provide the results of significance test for the following comparisons: (i) comparing the results of \sysGPT{} in 8-shots with (a) U-CoT+~\cite{pan2026readseeguidingunimodal}, and (b) the 8-shot results over \sysI{} and \sysPx{}. We also compare the 8-shot results of employed VLMs (i.e. \sysGPT{}, \sysI{}, and \sysPx{}) with their corresponding zero-shot counterparts. To perform the significance test, we employ the exact two-sided binomial \textsc{McNemar} test\footnote{\url{https://rpubs.com/mbh038/614538}} to assess significant changes in the model's classification. We note the $p$-values and present these results in Table~\ref{tab:significance_test}. As per the results obtained in the table, \ps{} (\sysGPT{} in an 8-shot setup) is significantly different from U-CoT+ as well as \sysI{} and \sysPx{}. Further, the 8-shot versions are significantly different from the zero-shot ones except for \sysPx{} for the MAMI dataset.


\begin{figure}
\centering
    \includegraphics[width=0.45\linewidth]{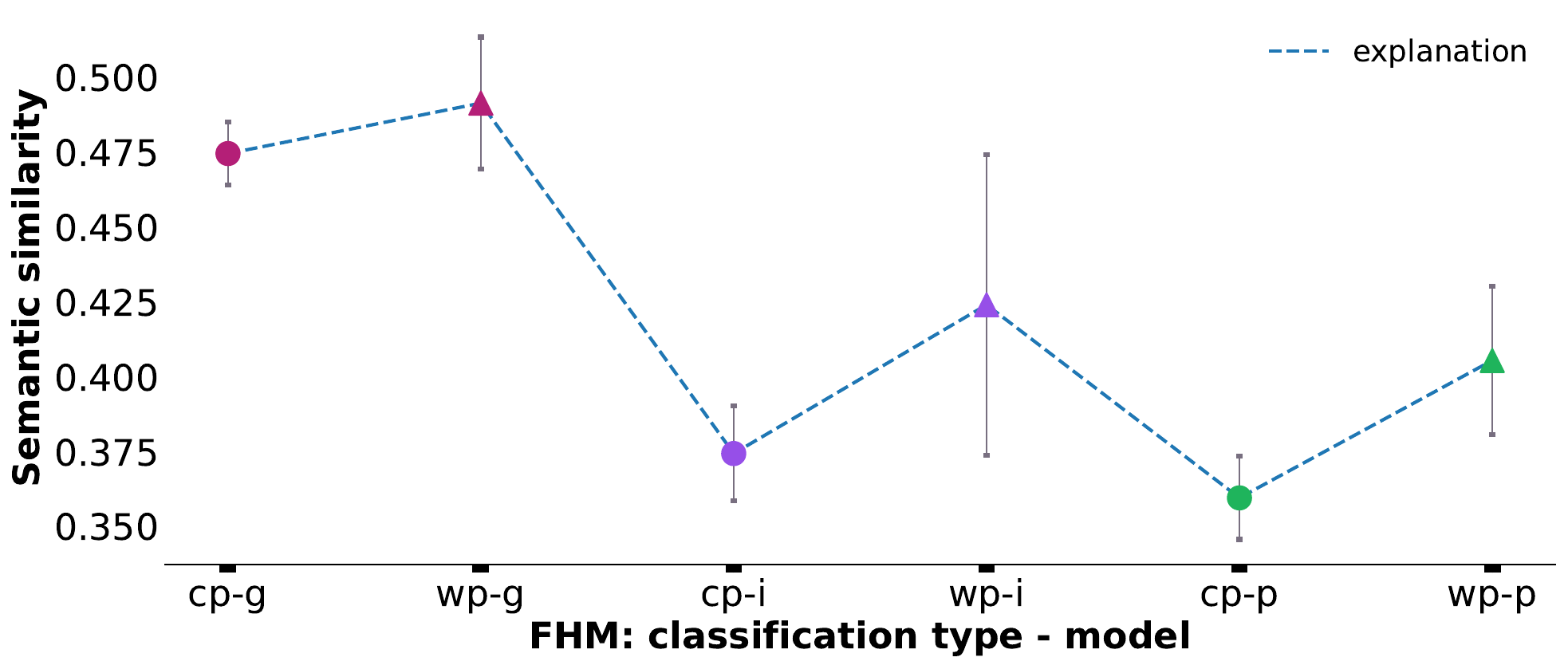}
    \includegraphics[width=0.45\linewidth]{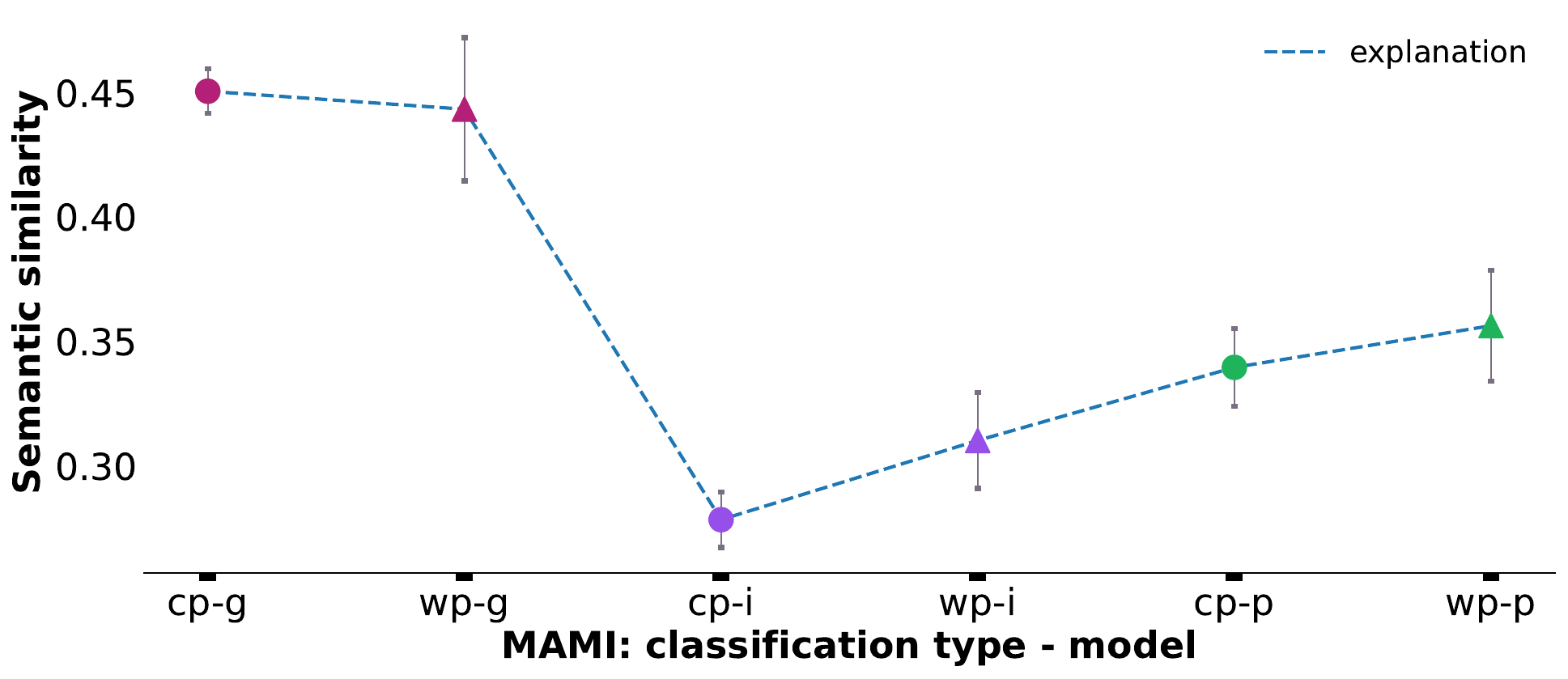}
\caption{Semantic relation based on semantic similarity for all considered models and datasets. Here, \texttt{cp}: correct positive, \texttt{wp}: wrong positive, \texttt{g}: \sysGPT{}, \texttt{i}: \sysI{}, and \texttt{p}: \sysPx{}.}
\label{fig:semantic_analysis}
\end{figure}

\section{Relationship between explanation and intervention} 
\textbf{Automatic score}: Measuring the semantic relationship of the generated explanation and intervention in case where the model predicted the input to be hateful (i.e., \texttt{cp} and \texttt{wp}) is important to understand if the latter follows from the former. To this purpose, we measure the semantic similarity (discussed in the results section of the main text) across all the models and datasets (Figure~\ref{fig:semantic_analysis}). \sysGPT{} has the highest semantic relationship over the other two models. Further, apart from the case of MAMI for \sysGPT{}, in all other cases, semantic similarity is higher for \texttt{wp} cases, signifying that the model strives to generate more appropriate explanation/intervention text to justify their wrong predictions.\\
\textbf{Human ratings}: As presented in the results section, we perform manual evaluation here as well on \sysGPT{} and \sysI{}. We randomly sample 200 samples and pass these to the two annotators. Following are the obtained results: (i) \sysGPT{}: FHM -- (0.54, 0.556) for \texttt{cp}, (0.56, 0.564) for \texttt{wp}, MAMI -- (0.616, 0.604) for \texttt{cp}, (0.548, 0.58) for \texttt{wp} and (ii) \sysI{} -- (0.556, 0.572) for \texttt{cp}, (0.52, 0.576) for \texttt{wp}, MAMI -- (0.592, 0.62) for \texttt{cp}, (0.536, 0.588) for \texttt{wp}. All these values in brackets again correspond to two different annotators and are similar to what we discussed in the results section. Overall these scores (and the observations therefore) match with the automated scores presented above.

\section{Word shift patterns}
\label{app:word_shift_patterns}
In the main content, we carried out extensive analysis to understand the textual properties of model-generated explanations and interventions.
In Figure~\ref{fig:word_shift_graph}, we further plot word shift graphs~\cite{Gallagher_2021}. Shannon Entropy shifts\footnote{\url{https://shifterator.readthedocs.io/en/latest/cookbook/frequency\_shifts.html\#shannon-entropy-shifts}} are employed since it identifies the surprisal appearance of a text in a corpus.
We compute two word shift graphs as follows: (i) compare the explanation of hateful vs non-hateful prediction for \sysGPT{}, and, (ii) compare the interventions of \sysGPT{} and \sysI{} (\sysI{} and \sysPx{} have similar patterns and hence one is shown for brevity).
We make the following observations: (i) explanations for predicting a data point as hateful contains words like `mocks', `stereotype',  `dehumanizing', and similar others that reflect negative sentiments; explanation for data points predicted as non-hateful focuses on words like `humorously',  `group', and `individual' that naturally reflect positive sentiments, (ii) the interventions of \sysGPT{} has words like `promote', `understanding', and  `avoid' that lean toward positivity as opposed to \sysI{} generated interventions that contain the words like `restricted', `inappropriate', and `offend'. As a conclusion, this word shift graph-based analysis strongly corroborates the sentiment analysis results presented in the main text.

\end{document}